\pdfoutput=1

\documentclass[11pt]{article}

\PassOptionsToPackage{table}{xcolor}

\usepackage[final]{acl}

\usepackage{times}
\usepackage{latexsym}

\usepackage[T1]{fontenc}

\usepackage[utf8]{inputenc}

\usepackage{multirow}   
\usepackage{tabularx}
\usepackage{booktabs}

\usepackage{graphicx} 
\usepackage{float}
\usepackage{amssymb}

\usepackage{hyperref}
\usepackage{xspace}

\usepackage{float}
\usepackage{amsmath}
\usepackage{bm}
\usepackage{algorithmicx}
\usepackage{algorithm}
\usepackage{algpseudocode}

\usepackage{amssymb}
\usepackage{pifont}
\usepackage{marvosym}
\newcommand{\cmark}{\ding{51}}%
\newcommand{\xmark}{\ding{55}}%
\newcommand{\gcmark}{\textcolor{green!25!black}{\cmark}}%
\usepackage{enumitem}

\usepackage{graphicx}
\usepackage{subcaption}
\newcommand{\successframe}{%
    \fcolorbox{green!55!black}{white}{\rule{0pt}{0.65em}\rule{0.65em}{0pt}}%
}
\newcommand{\failframe}{%
    \fcolorbox{orange!80!black}{white}{\rule{0pt}{0.65em}\rule{0.65em}{0pt}}%
}

\usepackage{microtype}
\usepackage[most]{tcolorbox}
\tcbset{
  colback=black!4,
  colframe=black!60,
  boxsep=1.2mm,
  left=1.2mm,right=1.2mm,top=1.2mm,bottom=1.2mm,
  title filled=true
}
\newcommand{\ours}{\textsc{GUI-CC}\xspace}

\newcommand{\nmodel}{12\xspace}
\newcommand{\nconfig}{18\xspace}

\newcommand{\huggingface}{\raisebox{-1.5pt}{\includegraphics[height=1.05em]{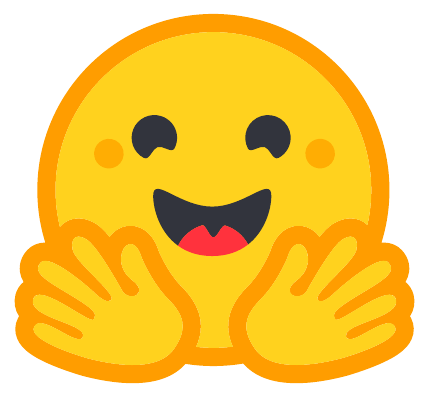}}\xspace}
\newcommand{\github}{\raisebox{-1.5pt}{\includegraphics[height=1.05em]{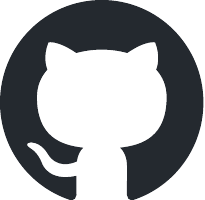}}\xspace}

\makeatletter
\newcommand\blfootnote[1]{%
  \begingroup\renewcommand\thefootnote{}\renewcommand\@makefnmark{}%
  \footnotetext{#1}\endgroup}
\makeatother

\title{\textsc{GUI-CC}: Benchmarking Contextual Consistency of GUI World Models as Agent Environments}

\author{
  Lin Fu\textsuperscript{1*} \quad Zheyuan Yang\textsuperscript{2*} \quad Tianhui Zhang\textsuperscript{3*} \quad Jinbiao Wei\textsuperscript{4} \\
  \bfseries Guo Gan\textsuperscript{1} \quad Boxu Liu\textsuperscript{5} \quad Yilun Zhao\textsuperscript{4\Letter} \quad Yu Rong\textsuperscript{6} \\[0.4em]
  \normalfont\normalsize
  \textsuperscript{1}Zhejiang University \quad
  \textsuperscript{2}Tongji University \quad
  \textsuperscript{3}University of California, San Diego \\
  \normalfont\normalsize
  \textsuperscript{4}Yale University \quad
  \textsuperscript{5}China University of Geosciences \quad
  \textsuperscript{6}DAMO Academy, Alibaba Group \\[0.3em]
  \normalfont\normalsize
  \github \href{https://github.com/Fu-Fu-Fu-Fu/GUI-CC}{Codebase} \quad
  \huggingface \href{https://huggingface.co/datasets/minuzero/GUI-CC}{Dataset}
}

\begin{document}
\maketitle
\blfootnote{\textsuperscript{*}Equal contributions.\quad\textsuperscript{\Letter}Corresponding author: \texttt{yilun.zhao@yale.edu}.}
\begin{abstract}

GUI world models are increasingly evaluated as one-step next-screen predictors, yet their intended use is often as multi-step environments for GUI agents. This mismatch leaves a key requirement under-tested: generated states must remain contextually consistent when they are repeatedly reused for future interaction.
We introduce \ours, a benchmark that evaluates contextual consistency of GUI world models as agent environments rather than isolated next-screen predictors.
\ours contains two complementary tracks: an offline reference-action track that rolls models along real mobile GUI trajectories, and an online agent-loop track that lets fixed probing agents interact with model-generated UIs.
We construct 500 offline trajectory tasks from GUIOdyssey and 200 emulator-verified online tasks across 30 mobile apps. \ours evaluates transition fidelity, transition plausibility, contextual consistency, and task progress. Experiments show that plausible single-step generation does not guarantee reliable environment simulation: current models often produce usable-looking screens while failing to preserve task-relevant context or support executable multi-step rollouts.
\end{abstract}

\section{Introduction}
\begin{figure*}[t]
    \centering
    \includegraphics[width=\textwidth]{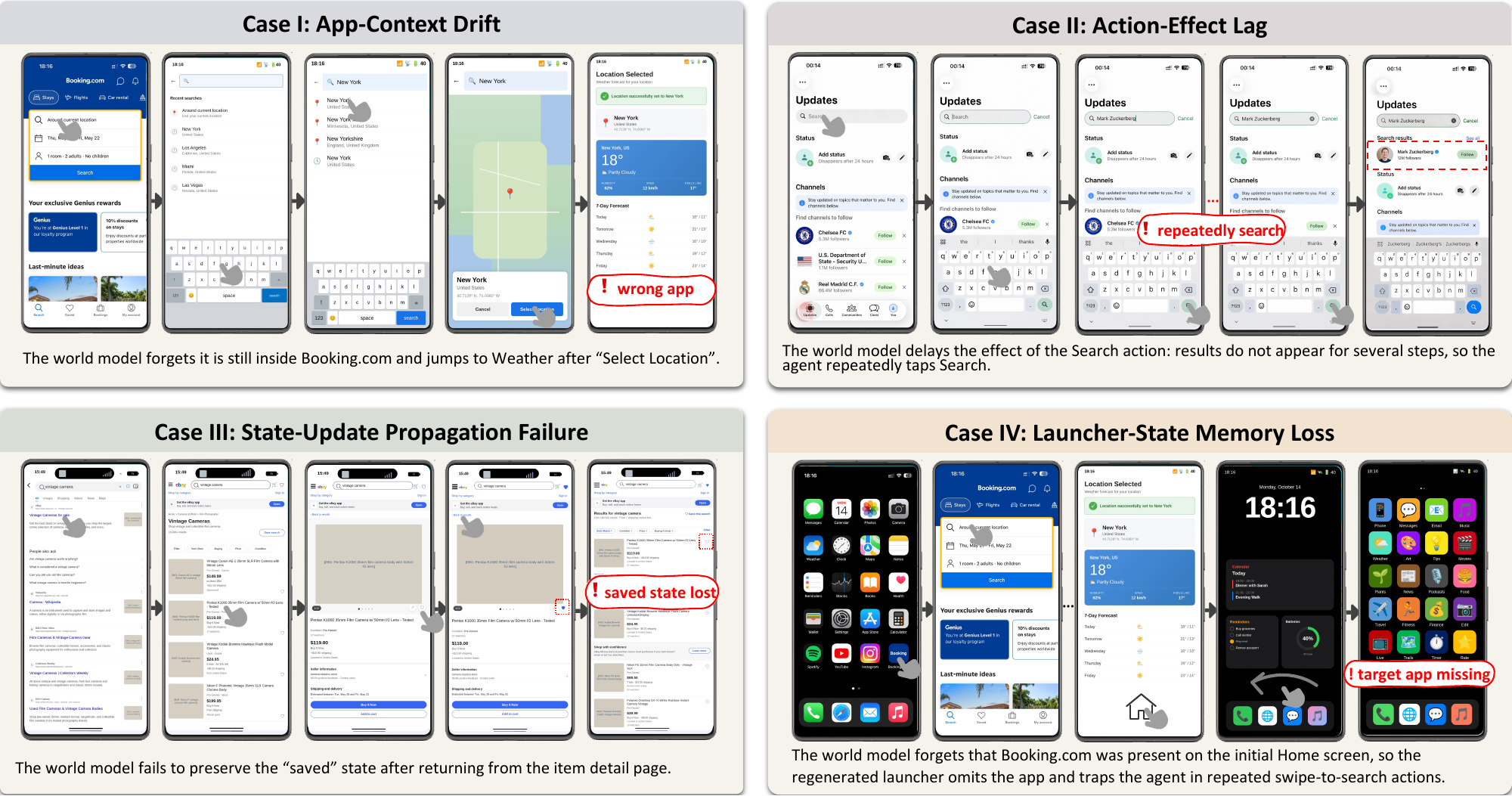}
    \caption{
    Representative failure cases of GUI world models under long-horizon rollouts.
    The examples illustrate app-context drift, action-effect lag, state-update propagation failure,
    and launcher-state memory loss. 
    }
    \label{fig:bad_case}
\end{figure*}

Autonomous GUI agents are defined by interactive execution: they observe graphical interfaces, issue UI-grounded actions, and complete tasks through environment state changes. Live web, desktop, and mobile benchmarks instantiate this setting with task initialization and execution-based success checks~\citep{webarena2023,osworld2024,androidworld2024}, while large-scale mobile datasets provide scalable logged demonstrations for training and evaluation~\citep{aitw2023,androidcontrol2024,guiodyssey2024}. Yet neither resource type alone provides cheap, reproducible counterfactual rollouts at scale: live environments are interactive but costly to reset and execute~\citep{gan2026androidcoachimproveonline,chen2025guishepherdreliableprocessreward}, while logged trajectories are scalable but fixed and cannot simulate unseen actions. GUI world models address this gap by learning action-conditioned transition dynamics, serving as surrogate environments for simulated rollouts, model comparison, and planning~\citep{wma2024,vimo2025,code2world2026,gworld2026,cuwm2026,xu2026mobileworldmodelguides}.

This surrogate-environment use changes what should be evaluated. A generated next screen may be locally plausible, but once it is fed back as the next state, future actions depend on whether app identity, navigation context, task-relevant entities, selected or created state, and action affordances remain coherent over the rollout. We call this requirement \emph{contextual consistency}. Prior GUI world-model evaluations~\citep{mobileworldbench2025,gworld2026,gebench2026,code2world2026} typically measure semantic or visual transition quality, scripted GUI generation, or isolated world-model outputs, but do not directly diagnose whether generated states remain coherent under repeated state reuse. As shown in \autoref{fig:bad_case}, visually plausible rollouts can still drift to the wrong app, delay action effects, lose saved state, or regenerate a launcher without the target app, leaving the agent in a plausible-looking but invalid environment.

We introduce \ours, a benchmark for evaluating \emph{contextual consistency} of GUI world models as agent environments. \ours contains two complementary tasks. The \emph{offline trajectory task} starts from a real initial UI and rolls out the world model along reference semantic action sequences, providing a controlled setting for testing whether generated states remain executable across real interaction trajectories. The \emph{online agent-loop task} lets fixed probing GUI agents act on model-generated UIs, directly testing whether the generated environment can support closed-loop task progress. We construct 500 offline trajectory tasks based on GUIOdyssey~\citep{guiodyssey2024} and 200 emulator-verified online agent-loop tasks over 30 mobile apps and 18 task templates. Our evaluation dimensions include transition fidelity, transition plausibility, contextual consistency, and task progress, separating local prediction quality from trajectory-level environment utility.

Our evaluation shows that plausible single-step prediction does not imply reliable rollout behavior. Models can obtain reasonable local transition scores while still losing task-relevant state, drifting across apps or pages, or failing to support later reference actions. History-conditioned inputs improve consistency in some cases, but the overall task-progress scores remain limited, especially in longer rollouts and closed-loop agent interaction. These results suggest that GUI world models are promising as learned rollout environments, but current systems still fall short of the contextual consistency needed for scalable agent training and evaluation.

\begin{table*}[t]
\centering
\small
\setlength{\tabcolsep}{3pt}
\renewcommand{\arraystretch}{1.15}
\resizebox{\linewidth}{!}{%
\begin{tabular}{l c l l c c c c c}
\toprule
\textbf{Evaluation} & \textbf{Release} & \textbf{Domain} & \textbf{Output State} & \textbf{Multi-Step} & \textbf{Pred.\ Fed Back} & \textbf{Agent-in-Loop} & \textbf{Direct WM Eval} & \textbf{Consistency Eval} \\
\midrule
WMA~\citep{wma2024}                                       & 2024-10 & Web     & Text            & \gcmark & \xmark & \xmark & \xmark & \xmark \\
MobileWorldBench~\citep{mobileworldbench2025}             & 2025-12 & Mobile  & Text   & \xmark & \xmark & \xmark & \gcmark & \xmark \\
gWorld~\citep{gworld2026}                                 & 2026-02 & Mobile  & HTML code       & \xmark & \xmark & \xmark & \gcmark & \xmark \\
GEBench~\citep{gebench2026}                               & 2026-02 & Mobile / desktop  & Image      & \gcmark & \xmark & \xmark & \gcmark & \xmark \\
Code2World~\citep{code2world2026}                         & 2026-02 & Mobile  & HTML code       & \xmark & \xmark & \xmark & \gcmark & \xmark \\
CUWM~\citep{cuwm2026}                                     & 2026-02 & Desktop & Text / image  & \xmark & \xmark & \xmark & \gcmark & \xmark \\
WebWorld~\citep{xiao2026webworldlargescaleworldmodel}     & 2026-02 & Web     & Structured      & \gcmark & \gcmark & \gcmark & \gcmark & \xmark \\
MobileWorld~\citep{xu2026mobileworldmodelguides}          & 2026-05 & Mobile  & Code / image / text   & \gcmark & \gcmark & \gcmark & \gcmark & \xmark \\
\midrule
\textbf{\ours Offline Track}    & 2026-05 & Mobile  & Code / image    & \gcmark & \gcmark & \xmark & \gcmark & \gcmark \\
\textbf{\ours Agent-Loop Track} & 2026-05 & Mobile  & Code / image    & \gcmark & \gcmark & \gcmark & \gcmark & \gcmark \\
\bottomrule
\end{tabular}%
}
\caption{
Comparison of \ours with prior GUI world-model evaluations.
\textbf{Pred.\ fed back}: predicted states are reused as the next environment state during the evaluation rollout.
\textbf{Agent-in-loop}: the rollout is driven by an agent acting on generated states.
\textbf{Direct WM eval}: the headline metric is computed on the world model's output rather than on downstream agent or search outcomes.
\textbf{Consistency eval}: the evaluation directly tests cross-step state-reuse consistency.
\ours measures reference-action executability and context persistence in the offline track, and milestone progress, loops, and stagnation in the agent-loop track.
GEBench reports an image-level temporal coherence score but does not test state-reuse consistency.
}
\label{tab:rw_comparison}
\end{table*}

\section{Related Work}

\paragraph{GUI Agent Benchmarks and Interaction Datasets.}
Live GUI agent benchmarks evaluate agents through execution in real web~\citep{webarena2023}, desktop~\citep{osworld2024}, and mobile~\citep{androidworld2024,DBLP:conf/acl/WeiZNC26,DBLP:journals/corr/abs-2604-27151,DBLP:journals/corr/abs-2605-19769,DBLP:journals/corr/abs-2606-24551,gan2026androidguiagentsrobust} applications, but their reliance on live execution makes large-scale rollout costly and hard to reproduce.
Large-scale mobile interaction datasets~\citep{aitw2023,androidcontrol2024,guiodyssey2024} provide instructions, screenshots, actions, and semantic annotations from real GUI trajectories, yet they are demonstrations rather than learned environments and do not by themselves test whether a model can replace the transition function during agent rollout.

\paragraph{GUI World Models.}
GUI world models predict the consequences of actions in graphical interfaces, letting agents reason about future UIs without executing every action on a live device.
Existing models differ mainly in how the predicted state is represented: textual or semantic transitions expressed in language~\citep{wma2024,mobileworldbench2025,mobiledreamer2026}, structured UI representations such as DOM or accessibility trees~\citep{xiao2026webworldlargescaleworldmodel}, image-level screenshot synthesis~\citep{vimo2025,xu2026mobileworldmodelguides}, renderable code that is rendered back into screenshots~\citep{code2world2026,gworld2026}, and hybrid pipelines combining textual transitions with visual realization~\citep{cuwm2026}.
Serving as an environment additionally requires generated states to remain mutually consistent when reused across the rollout.

\paragraph{Evaluating GUI World Models.}
Existing evaluations of GUI world models score one-step next-state fidelity inside model papers~\citep{wma2024,code2world2026,gworld2026,cuwm2026}, semantic next-state prediction in language~\citep{mobileworldbench2025}, multi-step GUI generation from scripted or instruction-conditioned inputs~\citep{gebench2026}, or downstream planning utility for an agent~\citep{xiao2026webworldlargescaleworldmodel}.
None of these settings tests whether predicted states remain a coherent environment when repeatedly fed back as the next input, where task-relevant latent state such as app identity, navigation history, created entities, selected options, and launcher layout must remain mutually consistent.
\ours fills this gap with an offline track that measures whether a real semantic action sequence remains executable under autoregressive prediction and an agent-loop track that measures whether fixed probing agents can make progress inside the generated environment.

\section{\ours Benchmark}
\label{sec:benchmark}

\begin{figure*}[t]
    \centering
    \begin{minipage}[t]{0.65\textwidth}
        \vspace{0pt}
        \centering
        \includegraphics[width=\linewidth]{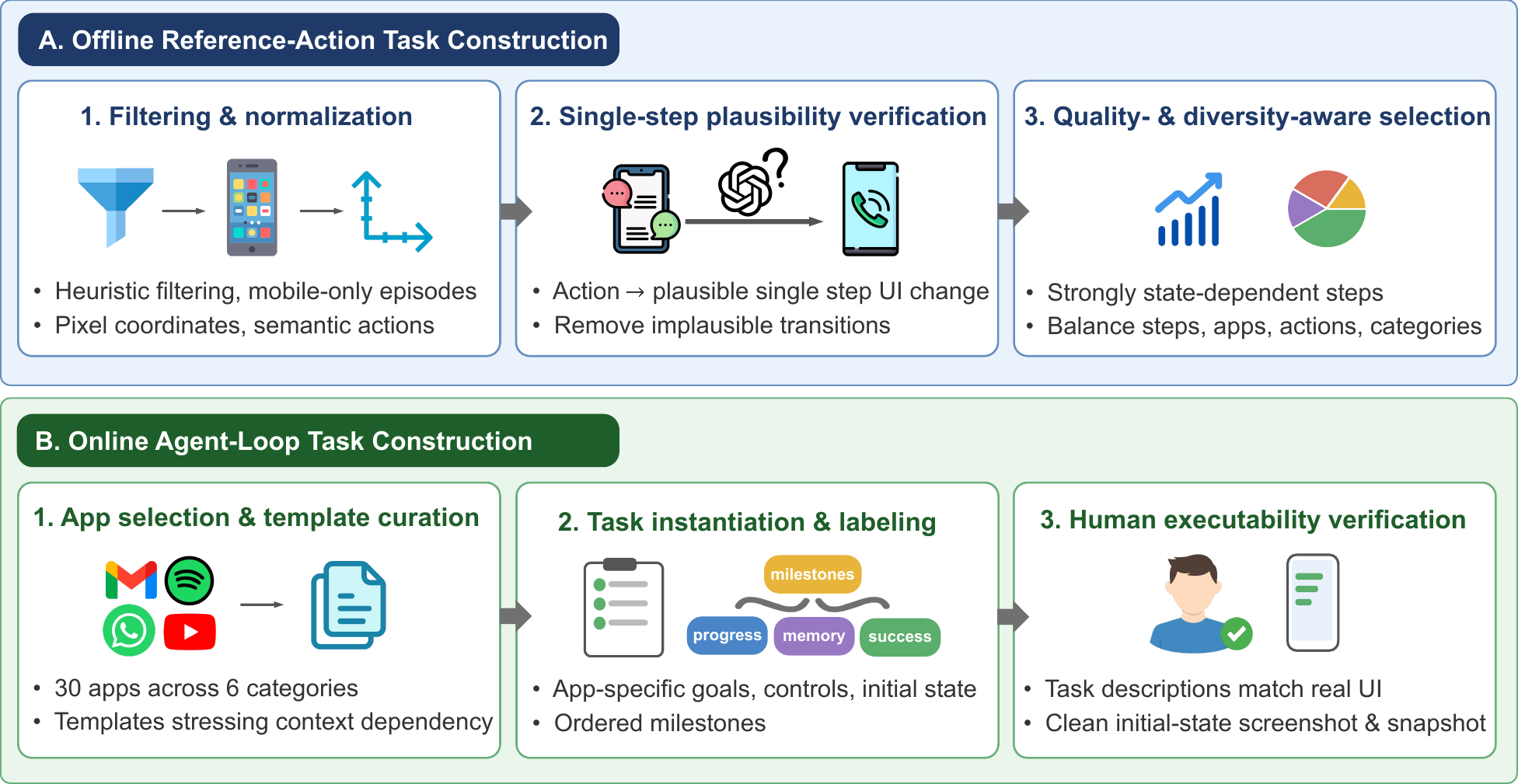}
    \end{minipage}%
    \hfill
    \begin{minipage}[t]{0.33\textwidth}
        \vspace*{20pt}
        \centering
        \resizebox{\linewidth}{!}{%
            \begin{tabular}{lrr}
            \toprule
            \textbf{Statistic} & \textbf{Offline} & \textbf{Online} \\
            \midrule
            \# Tasks & 500 & 200 \\
            \# Transitions & 4,905 & \textemdash{} \\
            Avg. steps / budget & 9.81 & 15.74 \\
            Min / max steps / budget & 7 / 14 & 6 / 24 \\
            \midrule
            \# Apps & 130 & 30 \\
            \# App categories & 6 & 6 \\
            \# Templates & \textemdash{} & 18 \\
            \midrule
            Avg. milestones & \textemdash{} & 4.0 \\
            Min / max milestones & \textemdash{} & 2 / 5 \\
            \bottomrule
            \end{tabular}%
        }
    \end{minipage}

    \caption{\textbf{(Left)} Data construction pipeline of GUI-CC. Both tasks are designed to stress contextual consistency under multi-step rollout. The offline trajectory task selects high-quality real mobile trajectories from GUIOdyssey, while the online agent-loop task creates emulator-verified tasks with ordered milestones for closed-loop agent evaluation. \textbf{(Right)} Statistics of the two tasks in \ours.
    Dashes indicate statistics that are not applicable to the corresponding track: the offline reference-action track does not use task templates or ordered milestones, while the online agent-loop track is evaluated with a step budget rather than a fixed set of reference transitions.
    }
    \label{fig:data_construct_and_stats}
\end{figure*}

This section presents \ours, a benchmark evaluating GUI world models as agent environments under multi-step action-conditioned state reuse; \autoref{fig:data_construct_and_stats} provides an overview of the data construction pipeline and track statistics.

\subsection{Problem Formulation}

\ours evaluates whether a GUI world model can serve as an interaction environment, not merely whether it can generate a plausible next screen.
A real GUI environment maintains a latent system state $s_t$ containing apps, the navigation stack, settings, user-created entities, and other hidden context.
At each step an agent chooses $a_t \sim \pi(\cdot \mid \tau_t, g)$ from history $\tau_t = (o_0, a_0, \ldots, o_t)$ and instruction $g$, and the environment transitions via
\[
(s_{t+1}, o_{t+1}) = E(s_t, a_t).
\]
A GUI world model replaces this transition with a learned function
\[
\hat{o}_{t+1} = W(H_t, a_t),
\]
where $H_t = (o_0, a_0, \hat{o}_1, \ldots, a_{t-1}, \hat{o}_t)$ is the generated history and each $\hat{o}_t$ becomes the input on which later actions are chosen or evaluated; $W$ never observes $s_t$, so whatever continuity the rollout exhibits must be reconstructed from the generated UIs.
A rollout is \emph{contextually consistent} if there exists a plausible latent-state sequence $(s_0, \ldots, s_T)$ under which each $\hat{o}_t$ is a rendering of $s_t$ and each action $a_t$ remains executable on $s_t$.
\ours operationalizes this property at the trajectory level: it measures whether the entire rollout admits such a latent-state explanation across multiple steps of action-conditioned reuse, rather than scoring isolated single-step predictions.

\subsection{Benchmark Tracks}
\label{sec:tracks}

\ours instantiates contextual consistency through two complementary tracks.
The \textbf{Offline Reference-Action Track} uses real GUI trajectories $(o_0, a_0, o_1, \ldots, a_{T-1}, o_T)$ as references: starting from $o_0$, the world model rolls out autoregressively along the reference semantic action sequence, and the reference screenshots are used only to score the predicted rollout, never as rollout inputs.
The \textbf{Online Agent-Loop Track} replaces the fixed action sequence with a frozen probing agent that chooses each $a_t$ on the previously generated rollout, until task completion, unusable output, or stagnation.
The two tracks share the autoregressive rollout structure but probe different aspects of environment consistency: the offline track fixes the action sequence to isolate the world model from agent behavior, while the online track lets the agent react to predictions to test whether the generated environment can sustain closed-loop interaction.

\subsection{Action Representation}
\label{sec:actions}

\ours represents every action in a semantic schema rather than as raw pixel coordinates.
Eight interaction types are shared by both tracks: \emph{tap} and \emph{long\_press} carry pixel coordinates and a natural-language target description; \emph{scroll} carries a direction and normalized start and end points; \emph{type\_text} carries the input text; \emph{open\_app} carries the target app name; and \emph{navigate\_back}, \emph{navigate\_home}, and \emph{wait} require no additional parameters.
The agent-loop track adds two terminal actions, \emph{terminate} and \emph{answer}; offline reference trajectories instantiate the five types that occur in GUIOdyssey (\emph{tap}, \emph{scroll}, \emph{navigate\_home}, \emph{navigate\_back}, \emph{long\_press}).
Coordinates are restored from GUIOdyssey's normalized $[0,1000]$ range to pixel coordinates using the screen resolution, and target descriptions are drawn from low-level instructions where available and labeled by a VLM otherwise.
This representation lets the offline track score reference-action executability based on whether the affordance remains visible in the predicted UI, decoupled from pixel-level reproduction.

\subsection{Offline Reference-Action Track}
\label{sec:offline_track}

The offline track is constructed from GUIOdyssey~\citep{guiodyssey2024}, a large-scale mobile GUI navigation dataset that provides screenshots, action records, and semantic annotations for multi-step cross-app interactions.

\paragraph{Stage 1: Filtering and Normalization.}
We first heuristically filter trajectories for world-model evaluation, keeping completed phone episodes with moderate steps and meaningful actions, while removing failed, repetitive, launcher-dominated, or structurally abnormal trajectories. We then normalize each retained trajectory into a canonical sequence $(o_0,a_0,o_1,\ldots,o_T)$ by restoring coordinates from the normalized $[0,1000]$ range to pixels and converting raw action records into semantic actions. This yields a unified trajectory format for autoregressive rollout and evaluation.

\paragraph{Stage 2: Single-Step Plausibility Verification.}
We verify normalized trajectories at the transition level to ensure each recorded action corresponds to a plausible single-step UI change. Although they come from real interactions, logs may contain missing operations or off-record state changes. We use GPT-5.5 to inspect each before-action-after triple and judge whether the after state can result from the recorded action. Trajectories with implausible transitions are removed, so the remaining references can serve as action-conditioned rollouts.

\paragraph{Stage 3: Quality- and Diversity-Aware Selection.}
From the verified pool, we select 500 offline trajectory tasks based on data quality and diversity. For data quality, we use GPT-5.5 to inspect the full trajectory and estimate the number of strongly state-dependent steps. We prioritize trajectories with more such steps, since they are more informative for evaluating contextual consistency. For diversity, we balance trajectory steps, apps, action types, task categories, and interaction patterns, ensuring the final set covers a broad range of rollout structures and GUI behaviors.

\subsection{Online Agent-Loop Track}
\label{sec:online_track}

The online track covers 30 mobile apps, spanning six categories: communication, travel \& navigation, shopping \& delivery, finance, media \& entertainment, and system \& utility.

\paragraph{Stage 1: App Selection and Task Template Curation.}
Annotators first brainstorm AI-assisted task templates, then manually review and refine them for executability and app-specific applicability. These templates stress contextual consistency through multi-step patterns, including cross-app navigation, state persistence after context changes, and revisiting task-relevant entities.

\paragraph{Stage 2: Task Instantiation and Labeling.}
Each selected template is instantiated by grounding it in the concrete UI flow of a target app. Annotators specify the task goal, app-specific pages or controls, required account or environment conditions, and initial-state description, turning a generic template into an executable task with a specific starting state and expected interaction path. Annotators also provide an ordered milestone sequence for evaluation. The milestones mark key states to be reached during execution, including intermediate progress, state persistence after context changes, and terminal task success. They allow the online task to measure whether the model-generated environment supports both task advancement and retention of established context.

\paragraph{Stage 3: Human Executability Verification.}
Finally, each online task is manually verified in an Android emulator. Annotators execute the task from the specified initial state, confirm that the instruction and milestones match the real UI, and save a clean initial-state screenshot and emulator snapshot for evaluation. Tasks with incorrect UI descriptions, unreachable milestones, or unstable conditions are revised or removed. The step budget is set from human completion length, typically around twice the number of human steps.

\subsection{Quality Control}

After construction, we perform an independent second-pass audit on both tracks. In the offline track, we re-check whether the screenshots, semantic actions, and transition sequence are coherent. In the online track, we re-check whether the instruction and initial state are clear, the milestones are observable, and the step budget is reasonable. Tasks with minor annotation issues are revised, while unreliable tasks are removed. This second-pass audit helps ensure that benchmark failures mainly reflect world-model inconsistency rather than data or task artifacts. For reproducibility, all VLM judges run with fixed model versions and frozen prompts, which we release together with the benchmark. More details of the data construction process are provided in Appendix~\ref{app:data_construction}.

\begin{table*}[t]
\centering
\setlength{\tabcolsep}{3.5pt}
\renewcommand{\arraystretch}{1.05}
\scriptsize

\resizebox{\textwidth}{!}{%
\begin{tabular}{@{}cll *{11}{>{\centering\arraybackslash}p{0.052\textwidth}}@{}}
\toprule
\multirow{2}{*}{\textbf{\#}} 
& \multirow{2}{*}{\textbf{World Model}} 
& \multirow{2}{*}{\textbf{Setting}} 
& \multicolumn{4}{c}{\textbf{Transition Fidelity}} 
& \multicolumn{3}{c}{\textbf{Transition Plausibility}} 
& \multicolumn{2}{c}{\textbf{CC}} 
& \multicolumn{1}{c}{\textbf{TP}} 
& \multirow{2}{*}{\textbf{Overall}} \\
\cmidrule(lr){4-7}
\cmidrule(lr){8-10}
\cmidrule(lr){11-12}
\cmidrule(lr){13-13}
& & 
& $\boldsymbol{S}_{\mathbf{ele}}$ & $\boldsymbol{S}_{\mathbf{lay}}$ & $\boldsymbol{S}_{\mathbf{sig}}$ & $\boldsymbol{S}_{\mathbf{dino}}$ 
& $\boldsymbol{S}_{\mathbf{ad}}$ & $\boldsymbol{S}_{\mathbf{id}}$ & $\boldsymbol{S}_{\mathbf{use}}$ 
& $\boldsymbol{S}_{\mathbf{cp}}$ & $\boldsymbol{S}_{\mathbf{rd}}$ 
& $\boldsymbol{S}_{\mathbf{rap}}$ 
& \\
\midrule

\multicolumn{13}{l}{\textbf{Code / HTML-output world models}} \\
\midrule

1  & Claude Opus 4.7     & w/ history  & \underline{16.8} & \underline{15.6} & \underline{73.4} & 47.1 & \textbf{69.5} & \textbf{79.0} & \textbf{99.6} & \textbf{65.0} & \textbf{93.8} & \textbf{16.4} & \textbf{57.6} \\
2  & GPT-5.5             & w/ history  & \textbf{17.8} & \textbf{16.6} & \textbf{75.0} & \textbf{49.8} & \underline{63.4} & 68.2 & 99.4 & \underline{59.5} & \underline{80.9} & \underline{15.3} & \underline{54.6} \\
3  & Claude Opus 4.7     & w/o history & 13.0 & 11.9 & 71.0 & 45.9 & 62.0 & \underline{78.3} & \underline{99.6} & 46.9 & 78.1 & 13.1 & 52.0 \\
4  & GPT-5.5             & w/o history & 13.1 & 11.8 & 71.5 & \underline{47.5} & 53.3 & 63.7 & 98.2 & 35.9 & 55.1 & 14.7 & 46.5 \\
5  & Code2World          & w/ history  & 9.8  & 8.3  & 68.9 & 36.0 & 49.1 & 61.5 & 96.0 & 38.0 & 58.9 & 9.5  & 43.6 \\
6  & MobileWorld-html-8B & w/ history & 10.3 & 9.0  & 70.3 & 40.6 & 47.6 & 60.2 & 93.4 & 30.7 & 53.8 & 8.0  & 42.4 \\
7  & MobileWorld-html-8B & w/o history  & 10.2 & 8.8  & 69.4 & 37.5 & 44.0 & 57.6 & 91.0 & 28.7 & 54.3 & 9.3  & 41.1 \\
8  & Code2World          & w/o history & 6.7  & 5.3  & 65.2 & 36.6 & 35.4 & 47.8 & 95.8 & 21.7 & 31.1 & 9.1  & 35.5 \\
9  & gWorld-32B          & w/ history  & 6.9  & 6.0  & 64.0 & 39.2 & 23.2 & 67.3 & 95.0 & 8.3  & 17.8 & 3.6  & 33.1 \\
10 & gWorld-8B           & w/ history  & 6.0  & 4.9  & 63.6 & 38.1 & 21.9 & 68.0 & 93.4 & 7.0  & 14.3 & 4.8  & 32.2 \\
11 & gWorld-32B          & w/o history & 5.0  & 4.2  & 62.6 & 38.4 & 19.5 & 66.0 & 91.9 & 5.0  & 17.0 & 3.5  & 31.3 \\
12 & gWorld-8B           & w/o history & 4.3  & 3.3  & 61.8 & 38.3 & 19.0 & 60.8 & 91.2 & 4.0  & 11.0 & 4.5  & 29.8 \\

\midrule

\multicolumn{13}{l}{\textbf{Direct image-generation world models}} \\
\midrule

1 & GPT Image 2            & w/o history & \textbf{16.9} & \textbf{14.7} & \textbf{77.3} & \textbf{40.0} & \underline{65.9} & \textbf{69.4} & \textbf{99.9} & \textbf{60.4} & \textbf{88.8} & \textbf{16.7} & \textbf{55.0} \\
2 & Gemini 3.1 Flash Image & w/o history & \underline{13.3} & \underline{11.6} & \underline{75.7} & \underline{39.8} & \textbf{69.6} & \underline{68.9} & 92.2 & \underline{57.6} & \underline{74.4} & \underline{14.0} & \underline{51.7} \\
3 & Vimo                   & w/o history & 4.8  & 3.2  & 65.3 & 28.3 & 41.0 & 38.9 & 39.0 & 8.4  & 20.4 & 4.3  & 25.4 \\
4 & Flux.2-dev             & w/o history & 9.4  & 8.9  & 69.9 & 28.1 & 16.4 & 4.1  & \underline{94.1} & 0.0  & 11.6 & 0.0  & 24.3 \\
5 & Qwen-Image-Edit-2511   & w/o history & 5.8  & 5.2  & 65.8 & 24.5 & 17.5 & 1.5  & 42.8 & 0.0  & 0.0  & 0.0  & 16.3 \\
6 & MobileWorld-Diffusion  & w/o history & 2.2  & 1.3  & 53.0 & 10.0 & 12.1 & 4.2  & 20.8 & 0.0  & 0.4  & 0.2  & 10.4 \\

\bottomrule
\end{tabular}%
}
\caption{
Offline reference-action track evaluation of GUI world models. 
The \emph{Overall} column is the 
unweighted mean across all ten metrics.
\emph{Setting} denotes the conditioning context: \emph{w/o history} uses only 
the current UI and action, while \emph{w/ history} uses multi-step historical states and actions.
\emph{CC} denotes \emph{Contextual Consistency}. \emph{TP} denotes \emph{Task Progress}.
The metrics evaluate UI element alignment ($S_{\mathrm{ele}}$), layout integrity ($S_{\mathrm{lay}}$), global similarity measured by SigLIP ($S_{\mathrm{sig}}$) and DINO ($S_{\mathrm{dino}}$), action adherence ($S_{\mathrm{ad}}$), action identifiability ($S_{\mathrm{id}}$), GUI state usability ($S_{\mathrm{use}}$), state and context persistence ($S_{\mathrm{cp}}$), action-controlled rollout dynamics ($S_{\mathrm{rd}}$), and reference-action progress ($S_{\mathrm{rap}}$).
Within each model group, the best score per column is \textbf{bold} and the second best is \underline{underlined}.
}
\label{tab:offline_wm_full}
\vspace{-0.5em}
\end{table*}

\begin{table*}[t]
\centering
\setlength{\tabcolsep}{3.5pt}
\renewcommand{\arraystretch}{1.05}
\scriptsize

\resizebox{\textwidth}{!}{%
\begin{tabular}{@{}cll *{7}{>{\centering\arraybackslash}p{0.075\textwidth}}@{}}
\toprule
\multirow{2}{*}{\textbf{\#}} 
& \multirow{2}{*}{\textbf{World Model}} 
& \multirow{2}{*}{\textbf{Setting}} 
& \multicolumn{3}{c}{\textbf{Transition Plausibility}} 
& \multicolumn{2}{c}{\textbf{CC}} 
& \multicolumn{1}{c}{\textbf{TP}} 
& \multirow{2}{*}{\textbf{Overall}} \\
\cmidrule(lr){4-6}
\cmidrule(lr){7-8}
\cmidrule(lr){9-9}
& & 
& $S_{\mathrm{ad}}$ & $S_{\mathrm{id}}$ & $S_{\mathrm{use}}$ 
& $S_{\mathrm{cp}}$ & $S_{\mathrm{rd}}$ 
& $S_{\mathrm{mp}}$ 
& \\

\midrule

\multicolumn{10}{l}{\textbf{Code / HTML-output world models}} \\
\midrule

1  & GPT-5.5             & w/ history  & \textbf{89.5} & \textbf{88.9} & \underline{99.9} & \textbf{93.3} & \textbf{99.2} & \textbf{70.9} & \textbf{90.3} \\
2  & GPT-5.5             & w/o history & 84.2 & 84.6 & \textbf{100.0} & 80.0 & \underline{95.0} & \underline{51.2} & \underline{82.5} \\
3  & Claude Opus 4.7     & w/ history  & \underline{84.7} & 84.0 & 99.4 & \underline{83.3} & 89.2 & 48.1 & 81.4 \\
4  & Claude Opus 4.7     & w/o history & 82.0 & \underline{88.0} & 99.4 & 57.5 & 78.3 & 41.4 & 74.4 \\
5  & Code2World          & w/ history  & 64.5 & 65.2 & 97.5 & 54.2 & 67.5 & 32.2 & 63.5 \\
6  & gWorld-32B          & w/ history  & 58.9 & 71.1 & 98.7 & 44.2 & 53.3 & 24.4 & 58.4 \\
7  & gWorld-8B           & w/ history  & 58.9 & 71.6 & 96.4 & 44.2 & 54.2 & 20.7 & 57.7 \\
8  & MobileWorld-html-8B & w/ history  & 61.6 & 64.1 & 91.7 & 44.2 & 54.2 & 19.9 & 56.0 \\
9  & Code2World          & w/o history & 61.1 & 60.5 & 96.1 & 38.3 & 54.2 & 19.4 & 54.9 \\
10 & MobileWorld-html-8B & w/o history & 60.6 & 66.1 & 95.5 & 26.7 & 50.0 & 22.6 & 53.6 \\
11 & gWorld-8B           & w/o history & 46.3 & 57.8 & 95.5 & 13.3 & 39.2 & 12.1 & 44.0 \\
12 & gWorld-32B          & w/o history & 42.5 & 63.7 & 97.4 & 11.7 & 31.7 & 15.4 & 43.7 \\

\midrule

\multicolumn{10}{l}{\textbf{Direct image-generation world models}} \\
\midrule

1 & Gemini 3.1 Flash Image & w/o history & \textbf{85.0} & \textbf{89.1} & \underline{95.6} & \textbf{70.8} & \underline{85.8} & \underline{43.7} & \textbf{78.3} \\
2 & GPT Image 2            & w/o history & \underline{82.6} & \underline{82.2} & \textbf{100.0} & \underline{68.3} & \textbf{90.8} & \textbf{43.9} & \underline{78.0} \\
3 & Vimo                   & w/o history & 37.8 & 40.2 & 41.6 & 15.8 & 20.0 & 14.0 & 28.2 \\
4 & Flux.2-dev             & w/o history & 13.3 & 3.8  & 92.1 & 2.5  & 21.7 & 1.0  & 22.4 \\
5 & Qwen-Image-Edit-2511   & w/o history & 14.4 & 4.9  & 42.9 & 2.5  & 0.8  & 1.4  & 11.2 \\
6 & MobileWorld-Diffusion  & w/o history & 12.2 & 2.3  & 31.2 & 0.0  & 0.0  & 3.3  & 8.2  \\

\bottomrule
\end{tabular}%
}
\caption{
Online agent-loop track evaluation of GUI world models. 
The \emph{Overall} column is the 
unweighted mean across all six metrics.
\emph{Setting} denotes the conditioning context: \emph{w/o history} uses only 
the current UI and action, while \emph{w/ history} uses multi-step historical states and actions.
\emph{CC} denotes \emph{Contextual Consistency}. \emph{TP} denotes \emph{Task Progress}.
The metrics evaluate action adherence ($S_{\mathrm{ad}}$), action identifiability ($S_{\mathrm{id}}$), GUI state usability ($S_{\mathrm{use}}$), state and context persistence ($S_{\mathrm{cp}}$), action-controlled rollout dynamics ($S_{\mathrm{rd}}$), and ordered milestone progress ($S_{\mathrm{mp}}$).
Within each model group, the best score per column is \textbf{bold} and the second best is \underline{underlined}.
}
\label{tab:online_wm_full}
\vspace{-0.5em}
\end{table*}

\subsection{Evaluation Metrics}
\label{sec:evaluation}

We evaluate GUI world models along four dimensions: \emph{transition fidelity}, \emph{transition plausibility}, \emph{contextual consistency}, and \emph{task progress}. The first two assess local prediction quality, extending Code2World~\citep{code2world2026} with GUI state usability; the latter two assess coherence and executability across multi-step rollouts.

\paragraph{Transition Fidelity.}
This dimension evaluates whether the predicted next UI matches the reference UI in visual content and structure.
SigLIP ($S_{\mathrm{sig}}$) and DINO ($S_{\mathrm{dino}}$) measure global similarity, while two fine-grained metrics assess UI details: (1) \emph{Element Alignment} ($S_{\mathrm{ele}}$) checks whether key UI elements are correctly present and aligned; (2) \emph{Layout Integrity} ($S_{\mathrm{lay}}$) checks whether the page layout and visual hierarchy are preserved.

\paragraph{Transition Plausibility.}
This dimension evaluates whether the predicted transition is a reasonable action-conditioned GUI update via three metrics: (1) \emph{Action Adherence} ($S_{\mathrm{ad}}$) checks whether the predicted UI reflects the intended effect of the action; (2) \emph{Action Identifiability} ($S_{\mathrm{id}}$) checks whether the action can be inferred from the visual change; (3) \emph{GUI State Usability} ($S_{\mathrm{use}}$) checks whether the predicted UI is readable, coherent, and usable for further interaction.

\paragraph{Contextual Consistency.}
This dimension evaluates whether rollout remains consistent with history $H_t$ via two metrics: (1) \emph{State and Context Persistence} ($S_{\mathrm{cp}}$) checks whether task-relevant states, entities, pages, and targets persist appropriately; (2) \emph{Action-Controlled Rollout Dynamics} ($S_{\mathrm{rd}}$) checks whether the rollout follows action sequence without drift, freezing, loops, or layout collapse.

\paragraph{Task Progress.}
This dimension evaluates whether generated rollout supports multi-step task execution with two metrics: (1) \emph{Reference Action Progress} ($S_{\mathrm{rap}}$) measures how many reference actions are consecutively supported in the offline track; (2) \emph{Ordered Milestone Progress} ($S_{\mathrm{mp}}$) measures the ordered prefix of task milestones reached in the online track.

\paragraph{Scoring and Aggregation.}
For VLM-judged metrics, each metric is mapped to $[0,1]$ under its own rubric: $S_{\mathrm{ele}}$ and $S_{\mathrm{lay}}$ are 1--10 ratings mapped by $(x-1)/9$, $S_{\mathrm{ad}}$ is a 0--10 rating divided by 10, $S_{\mathrm{id}}$ is a binary action-category match, and $S_{\mathrm{use}}$, $S_{\mathrm{cp}}$, and $S_{\mathrm{rd}}$ each average five binary criteria; encoder-based metrics are cosine similarities clamped to $[0,1]$.
Step-level metrics are aggregated by averaging over all transitions in an episode and then over episodes, $S_m = \frac{1}{|\mathcal{E}|} \sum_{e \in \mathcal{E}} \frac{1}{T_e} \sum_{t=1}^{T_e} S_m(e, t)$, while trajectory-level metrics are scored once per episode; in particular, $S_{\mathrm{rap}}(e) = L_e / T_e$, where $L_e$ is the longest reference-action prefix the predicted rollout supports.
Detailed metric definitions, prompts, and scoring rubrics are provided in Appendix~\ref{app:eval_metrics}.

\section{Experiments}

\subsection{Experimental Setup}
\label{sec:exp_setup}

\paragraph{Evaluated models.}
We evaluate \nmodel GUI world models across two output modalities, yielding \nconfig evaluated configurations per track. The first group comprises code-based models that generate renderable HTML, covering both proprietary general models adapted with HTML-generation prompts and specialized open-source GUI world models~\citep{code2world2026,xu2026mobileworldmodelguides,gworld2026}. Their outputs are rendered into screenshots before evaluation. The second group comprises direct image-generation models~\citep{openai2026gptimage2,google2026gemini31image,wu2025qwenimagetechnicalreport,flux-2-2025,vimo2025,xu2026mobileworldmodelguides}. \mbox{Appendix~\ref{app:model_list}} lists model specifications.

\paragraph{Evaluation protocol.}
All models are evaluated as action-conditioned transition predictors. Given the current UI and a semantic action, each model predicts the next UI state, either as renderable HTML or as an image. For models that can consume rollout history, we additionally report a \emph{w/ history} variant that prepends up to three previous observation-action pairs; this is reported separately from the native \emph{w/o history} setting.
For the online agent-loop task, we use GPT-5.5 as the fixed probing GUI agent across all world models, so performance differences primarily reflect the generated environments rather than different agent policies. For VLM-judged metrics, we also use GPT-5.5 as a frozen judge with fixed prompts and rubrics. Transition fidelity is reported only for the offline trajectory task, where reference screenshots are available. Transition plausibility and contextual consistency are reported for both tasks. Task progress uses reference-action progress offline and ordered milestone progress online. Scores are task averages on a $0$--$100$ scale.

\subsection{Main Results}

Tables~\ref{tab:offline_wm_full} and~\ref{tab:online_wm_full} report offline and online results.

\paragraph{\ours is challenging for current GUI world models.}
Task progress is the most difficult dimension across both tracks. In the offline track, even the best model reaches only 16.7 reference-action progress, showing that most rollouts quickly become unable to support later reference actions. In the online track, GPT-5.5 achieves the strongest milestone progress, while specialized GUI world models remain substantially lower.

\paragraph{Global similarity is easier than fine-grained UI fidelity.}
Offline transition-fidelity metrics show a clear gap between coarse visual similarity and detailed UI reconstruction. Strong models obtain relatively high embedding-based similarity scores, but fine-grained element alignment and layout integrity remain low, with the best scores below 20. Thus, models often preserve rough screen semantics while misplacing interaction-critical elements, text, controls, and layouts.

\paragraph{Usable-looking GUIs do not imply correct dynamics.}
Transition-plausibility metrics separate visual usability from action-conditioned environment dynamics. Many models generate readable GUI-like screens, and several strong models obtain near-perfect usability scores. However, high usability can coexist with poor action grounding and low task progress. Flux.2-dev is a representative case: it achieves high usability in both tracks, but obtains near-zero task progress, indicating that it often produces plausible interfaces without modeling the actual consequence of the action. This suggests that plausible GUI images may still be poor agent environments if they do not reflect the action or preserve the correct app/page state.

\paragraph{History helps consistency but not enough task progress.}
Adding observation-action history generally improves models that can use it, especially on contextual-consistency metrics. For example, GPT-5.5 improves from 82.5 to 90.3 overall in the online track, and Code2World improves in both offline and online settings. However, these gains do not translate proportionally into task progress. For Code2World, offline state/context persistence and rollout dynamics improve substantially, but reference-action progress changes only from 9.1 to 9.5. This suggests that history helps surface-level continuity, but does not fully solve long-horizon state maintenance or action-effect propagation.

\subsection{Analysis}
\label{sec:analysis}

\paragraph{Automatic metrics are useful but incomplete.}
Automatic metrics provide scalable diagnostics, but they do not fully replace human judgment of multi-step consistency. History conditioning illustrates this limitation: some models obtain much higher transition and contextual-consistency scores after adding history, while task progress improves only marginally. Manual inspection shows that many rollouts with high CC scores still contain fine-grained task-state errors, such as selecting the wrong item, keeping a stale query, failing to propagate a saved state, or staying on a visually plausible but task-incorrect page. These errors directly affect whether later actions remain executable, but VLM judges may under-penalize them when local visual continuity and layout appear reasonable.
Transition fidelity metrics have a related limitation. Global embedding similarity captures coarse semantic resemblance, while element alignment and layout integrity compare against a single ground-truth next UI. Low scores may reflect real element or layout failures, but may also penalize a different yet plausible and executable next state. We therefore treat automatic metrics as scalable diagnostic signals rather than complete measures of GUI world-model quality, especially when multiple valid UI transitions may exist.

\paragraph{Error analysis.}
To understand failure sources, we randomly sample 200 failed rollouts from GPT-5.5 and GPT Image 2 and manually annotate the primary failure type. We identify three categories. \emph{Missing world knowledge} accounts for about 42\% of failures, where the model lacks app-specific, Android-level, or common UI transition knowledge, such as the correct launch state after tapping an app icon or the expected behavior of search bars, dialogs, save buttons, text inputs, and list pages.
\emph{Error accumulation} accounts for about 33\% of failures. A small early deviation is reused as the next state and amplifies into repeated searches, frozen screens, layout collapse, incorrect navigation, or divergence from the reference trajectory. This failure mode is exposed by autoregressive rollout but hidden by teacher-forced evaluation.
\emph{Context inconsistency} accounts for about 25\% of failures, including lost app identity, corrupted navigation history, forgotten typed text or queries, missing saved/bookmarked states, disappeared selected objects, and resampled launcher layouts. These errors may still yield readable screens, but break the assumption that the rollout remains in the same coherent world.
These categories are often coupled: missing transition knowledge can trigger the first error, autoregressive reuse amplifies it, and the rollout eventually becomes contextually inconsistent or task-inexecutable. Improving GUI world models therefore requires not only stronger visual generation, but also app-specific transition knowledge, long-term state maintenance, and stable action-effect propagation. Detailed case studies are provided in Appendix~\ref{app:case_study}.

\section{Conclusion}

We presented \ours, a benchmark that evaluates GUI world models beyond isolated next-screen fidelity, emphasizing contextual consistency and task completion across interactions. Its two-track design keeps the central environment question explicit: the agent-loop track tests whether model-generated GUIs remain usable by probing agents, while the offline trajectory track provides scalable multi-step diagnosis. Our results show that current GUI world models capture parts of local action dynamics, but still struggle to preserve task-relevant state through autoregressive rollout. High visual plausibility and contextual-consistency scores do not reliably translate into task progress, exposing a gap between convincing screens and executable environments. The limited gains from short histories further suggest that future models need persistent state representations and stronger action-effect knowledge rather than visual memory alone. Closing this gap is essential for scalable GUI agents.

\section*{Limitations}

GUI-CC is intended as an initial step toward evaluating GUI world models as reusable agent environments. The current benchmark focuses on mobile GUI tasks, where screenshots, action semantics, and emulator verification can be standardized at scale. A natural next direction is to extend the same contextual-consistency framework to web, desktop, and multi-device environments, where state persistence may involve richer browser sessions, file systems, or cross-device context. Another direction is to broaden the online agent-loop track beyond a single fixed probing agent. While using a fixed agent makes world-model comparisons controlled, evaluating with agents that differ in planning strength, recovery strategy, and prompting style would provide a more complete picture of how generated environments support diverse interaction policies.

Our evaluation also points to opportunities for improving consistency assessment itself. VLM-based judges provide scalable diagnostic signals, but future versions of GUI-CC can incorporate more human-validated checks, structured UI-state probes, and executable verification for fine-grained task states such as selected items, saved entities, stale queries, and persistent settings. Finally, GUI world models differ in their native input and output interfaces. Future benchmark releases can support richer model-specific interfaces, including structured state representations, memory modules, simulator APIs, and hybrid code-image outputs, while retaining a common evaluation protocol for rollout consistency and task progress.

\bibliography{custom}

@misc{xiao2026webworldlargescaleworldmodel,
      title={WebWorld: A Large-Scale World Model for Web Agent Training}, 
      author={Zikai Xiao and Jianhong Tu and Chuhang Zou and Yuxin Zuo and Zhi Li and Peng Wang and Bowen Yu and Fei Huang and Junyang Lin and Zuozhu Liu},
      year={2026},
      eprint={2602.14721},
      archivePrefix={arXiv},
      primaryClass={cs.AI},
      url={https://arxiv.org/abs/2602.14721}, 
}

@misc{webarena2023,
      title={WebArena: A Realistic Web Environment for Building Autonomous Agents}, 
      author={Shuyan Zhou and Frank F. Xu and Hao Zhu and Xuhui Zhou and Robert Lo and Abishek Sridhar and Xianyi Cheng and Tianyue Ou and Yonatan Bisk and Daniel Fried and Uri Alon and Graham Neubig},
      year={2024},
      eprint={2307.13854},
      archivePrefix={arXiv},
      primaryClass={cs.AI},
      url={https://arxiv.org/abs/2307.13854}, 
}

@misc{osworld2024,
      title={OSWorld: Benchmarking Multimodal Agents for Open-Ended Tasks in Real Computer Environments}, 
      author={Tianbao Xie and Danyang Zhang and Jixuan Chen and Xiaochuan Li and Siheng Zhao and Ruisheng Cao and Toh Jing Hua and Zhoujun Cheng and Dongchan Shin and Fangyu Lei and Yitao Liu and Yiheng Xu and Shuyan Zhou and Silvio Savarese and Caiming Xiong and Victor Zhong and Tao Yu},
      year={2024},
      eprint={2404.07972},
      archivePrefix={arXiv},
      primaryClass={cs.AI},
      url={https://arxiv.org/abs/2404.07972}, 
}

@misc{androidworld2024,
      title={AndroidWorld: A Dynamic Benchmarking Environment for Autonomous Agents}, 
      author={Christopher Rawles and Sarah Clinckemaillie and Yifan Chang and Jonathan Waltz and Gabrielle Lau and Marybeth Fair and Alice Li and William Bishop and Wei Li and Folawiyo Campbell-Ajala and Daniel Toyama and Robert Berry and Divya Tyamagundlu and Timothy Lillicrap and Oriana Riva},
      year={2025},
      eprint={2405.14573},
      archivePrefix={arXiv},
      primaryClass={cs.AI},
      url={https://arxiv.org/abs/2405.14573}, 
}

@misc{aitw2023,
      title={Android in the Wild: A Large-Scale Dataset for Android Device Control}, 
      author={Christopher Rawles and Alice Li and Daniel Rodriguez and Oriana Riva and Timothy Lillicrap},
      year={2023},
      eprint={2307.10088},
      archivePrefix={arXiv},
      primaryClass={cs.LG},
      url={https://arxiv.org/abs/2307.10088}, 
}

@misc{androidcontrol2024,
      title={On the Effects of Data Scale on UI Control Agents}, 
      author={Wei Li and William Bishop and Alice Li and Chris Rawles and Folawiyo Campbell-Ajala and Divya Tyamagundlu and Oriana Riva},
      year={2024},
      eprint={2406.03679},
      archivePrefix={arXiv},
      primaryClass={cs.AI},
      url={https://arxiv.org/abs/2406.03679}, 
}

@inproceedings{guiodyssey2024,
  title={GUIOdyssey: A comprehensive dataset for cross-app GUI navigation on mobile devices},
  author={Lu, Quanfeng and Shao, Wenqi and Liu, Zitao and Du, Lingxiao and Meng, Fanqing and Li, Boxuan and Chen, Botong and Huang, Siyuan and Zhang, Kaipeng and Luo, Ping},
  booktitle={Proceedings of the IEEE/CVF International Conference on Computer Vision},
  pages={22404--22414},
  year={2025}
}

@misc{vimo2025,
      title={ViMo: A Generative Visual GUI World Model for App Agents}, 
      author={Dezhao Luo and Bohan Tang and Kang Li and Georgios Papoudakis and Jifei Song and Shaogang Gong and Jianye Hao and Jun Wang and Kun Shao},
      year={2025},
      eprint={2504.13936},
      archivePrefix={arXiv},
      primaryClass={cs.HC},
      url={https://arxiv.org/abs/2504.13936}, 
}

@misc{code2world2026,
      title={Code2World: A GUI World Model via Renderable Code Generation}, 
      author={Yuhao Zheng and Li'an Zhong and Yi Wang and Rui Dai and Kaikui Liu and Xiangxiang Chu and Linyuan Lv and Philip Torr and Kevin Qinghong Lin},
      year={2026},
      eprint={2602.09856},
      archivePrefix={arXiv},
      primaryClass={cs.CV},
      url={https://arxiv.org/abs/2602.09856}, 
}

@misc{gworld2026,
      title={Generative Visual Code Mobile World Models}, 
      author={Woosung Koh and Sungjun Han and Segyu Lee and Se-Young Yun and Jamin Shin},
      year={2026},
      eprint={2602.01576},
      archivePrefix={arXiv},
      primaryClass={cs.LG},
      url={https://arxiv.org/abs/2602.01576}, 
}

@misc{cuwm2026,
      title={Computer-Using World Model}, 
      author={Yiming Guan and Rui Yu and John Zhang and Lu Wang and Chaoyun Zhang and Liqun Li and Bo Qiao and Si Qin and He Huang and Fangkai Yang and Pu Zhao and Lukas Wutschitz and Samuel Kessler and Huseyin A Inan and Robert Sim and Saravan Rajmohan and Qingwei Lin and Dongmei Zhang},
      year={2026},
      eprint={2602.17365},
      archivePrefix={arXiv},
      primaryClass={cs.SE},
      url={https://arxiv.org/abs/2602.17365}, 
}

@misc{gebench2026,
      title={GEBench: Benchmarking Image Generation Models as GUI Environments}, 
      author={Haodong Li and Jingwei Wu and Quan Sun and Guopeng Li and Juanxi Tian and Huanyu Zhang and Yanlin Lai and Ruichuan An and Hongbo Peng and Yuhong Dai and Chenxi Li and Chunmei Qing and Jia Wang and Ziyang Meng and Zheng Ge and Xiangyu Zhang and Daxin Jiang},
      year={2026},
      eprint={2602.09007},
      archivePrefix={arXiv},
      primaryClass={cs.AI},
      url={https://arxiv.org/abs/2602.09007}, 
}

@misc{mobileworldbench2025,
      title={MobileWorldBench: Towards Semantic World Modeling For Mobile Agents}, 
      author={Shufan Li and Konstantinos Kallidromitis and Akash Gokul and Yusuke Kato and Kazuki Kozuka and Aditya Grover},
      year={2025},
      eprint={2512.14014},
      archivePrefix={arXiv},
      primaryClass={cs.AI},
      url={https://arxiv.org/abs/2512.14014}, 
}

@misc{wma2024,
      title={Web Agents with World Models: Learning and Leveraging Environment Dynamics in Web Navigation}, 
      author={Hyungjoo Chae and Namyoung Kim and Kai Tzu-iunn Ong and Minju Gwak and Gwanwoo Song and Jihoon Kim and Sunghwan Kim and Dongha Lee and Jinyoung Yeo},
      year={2025},
      eprint={2410.13232},
      archivePrefix={arXiv},
      primaryClass={cs.CL},
      url={https://arxiv.org/abs/2410.13232}, 
}

@misc{mobiledreamer2026,
      title={MobileDreamer: Generative Sketch World Model for GUI Agent}, 
      author={Yilin Cao and Yufeng Zhong and Zhixiong Zeng and Liming Zheng and Jing Huang and Haibo Qiu and Peng Shi and Wenji Mao and Wan Guanglu},
      year={2026},
      eprint={2601.04035},
      archivePrefix={arXiv},
      primaryClass={cs.AI},
      url={https://arxiv.org/abs/2601.04035}, 
}

@misc{xu2026mobileworldmodelguides,
      title={How Mobile World Model Guides GUI Agents?}, 
      author={Weikai Xu and Kun Huang and Yunren Feng and Jiaxing Li and Yuhan Chen and Yuxuan Liu and Zhizheng Jiang and Heng Qu and Pengzhi Gao and Wei Liu and Jian Luan and Xiaolin Hu and Bo An},
      year={2026},
      eprint={2605.10347},
      archivePrefix={arXiv},
      primaryClass={cs.AI},
      url={https://arxiv.org/abs/2605.10347}, 
}

@misc{openai2026gpt55,
  title={{GPT-5.5} {System Card}},
  author={{OpenAI}},
  year={2026},
  note={\url{https://openai.com/index/gpt-5-5-system-card/}},
}

@misc{anthropic2026opus47,
  title={Claude {Opus 4.7} {System Card}},
  author={{Anthropic}},
  year={2026},
  note={\url{https://anthropic.com/claude-opus-4-7-system-card}},
}

@misc{wu2025qwenimagetechnicalreport,
      title={Qwen-Image Technical Report}, 
      author={Chenfei Wu and Jiahao Li and Jingren Zhou and Junyang Lin and Kaiyuan Gao and Kun Yan and Sheng-ming Yin and Shuai Bai and Xiao Xu and Yilei Chen and Yuxiang Chen and Zecheng Tang and Zekai Zhang and Zhengyi Wang and An Yang and Bowen Yu and Chen Cheng and Dayiheng Liu and Deqing Li and Hang Zhang and Hao Meng and Hu Wei and Jingyuan Ni and Kai Chen and Kuan Cao and Liang Peng and Lin Qu and Minggang Wu and Peng Wang and Shuting Yu and Tingkun Wen and Wensen Feng and Xiaoxiao Xu and Yi Wang and Yichang Zhang and Yongqiang Zhu and Yujia Wu and Yuxuan Cai and Zenan Liu},
      year={2025},
      eprint={2508.02324},
      archivePrefix={arXiv},
      primaryClass={cs.CV},
      url={https://arxiv.org/abs/2508.02324}, 
}

@misc{flux-2-2025,
    author={Black Forest Labs},
    title={{FLUX.2: Frontier Visual Intelligence}},
    year={2025},
    howpublished={\url{https://bfl.ai/blog/flux-2}},
}

@misc{openai2026gptimage2,
  title={Introducing {ChatGPT} {Images 2.0}},
  author={{OpenAI}},
  year={2026},
  note= {\url{https://openai.com/index/introducing-chatgpt-images-2-0/}},
}

@misc{google2026gemini31image,
  title={{Gemini 3.1 Flash Image Model Card}},
  author={{Google DeepMind}},
  year={2026},
  note={\url{https://deepmind.google/models/model-cards/gemini-3-1-flash-image/}},
}

@misc{gan2026androidcoachimproveonline,
      title={Android Coach: Improve Online Agentic Training Efficiency with Single State Multiple Actions}, 
      author={Guo Gan and Yuxuan Ding and Cong Chen and Yuwei Ren and Yin Huang and Hong Zhou},
      year={2026},
      eprint={2604.07277},
      archivePrefix={arXiv},
      primaryClass={cs.LG},
      url={https://arxiv.org/abs/2604.07277}, 
}

@misc{chen2025guishepherdreliableprocessreward,
      title={GUI-Shepherd: Reliable Process Reward and Verification for Long-Sequence GUI Tasks}, 
      author={Cong Chen and Kaixiang Ji and Hao Zhong and Muzhi Zhu and Anzhou Li and Guo Gan and Ziyuan Huang and Cheng Zou and Jiajia Liu and Jingdong Chen and Hao Chen and Chunhua Shen},
      year={2025},
      eprint={2509.23738},
      archivePrefix={arXiv},
      primaryClass={cs.AI},
      url={https://arxiv.org/abs/2509.23738}, 
}

@inproceedings{DBLP:conf/acl/WeiZNC26,
  author       = {Jinbiao Wei and
                  Yilun Zhao and
                  Kangqi Ni and
                  Arman Cohan},
  editor       = {Maria Liakata and
                  Viviane P. Moreira and
                  Jiajun Zhang and
                  David Jurgens},
  title        = {Anchor: Branch-Point Data Generation for {GUI} Agents},
  booktitle    = {Proceedings of the 64th Annual Meeting of the Association for Computational
                  Linguistics (Volume 1: Long Papers), {ACL} 2026, San Diego, California,
                  United States, July 2-7, 2026},
  pages        = {17031--17047},
  publisher    = {Association for Computational Linguistics},
  year         = {2026},
  url          = {https://doi.org/10.18653/v1/2026.acl-long.774},
  doi          = {10.18653/V1/2026.ACL-LONG.774},
  bibsource    = {dblp computer science bibliography, https://dblp.org}
}

@article{DBLP:journals/corr/abs-2604-27151,
  author       = {Jinbiao Wei and
                  Kangqi Ni and
                  Yilun Zhao and
                  Guo Gan and
                  Arman Cohan},
  title        = {Step-level Optimization for Efficient Computer-use Agents},
  journal      = {CoRR},
  volume       = {abs/2604.27151},
  year         = {2026},
  url          = {https://doi.org/10.48550/arXiv.2604.27151},
  doi          = {10.48550/ARXIV.2604.27151},
  eprinttype   = {arXiv},
  eprint       = {2604.27151},
  bibsource    = {dblp computer science bibliography, https://dblp.org}
}

@article{DBLP:journals/corr/abs-2605-19769,
  author       = {Jinbiao Wei and
                  Qianran Ma and
                  Yilun Zhao and
                  Xiao Zhou and
                  Kangqi Ni and
                  Guo Gan and
                  Arman Cohan},
  title        = {OpenComputer: Verifiable Software Worlds for Computer-Use Agents},
  journal      = {CoRR},
  volume       = {abs/2605.19769},
  year         = {2026},
  url          = {https://doi.org/10.48550/arXiv.2605.19769},
  doi          = {10.48550/ARXIV.2605.19769},
  eprinttype   = {arXiv},
  eprint       = {2605.19769},
  bibsource    = {dblp computer science bibliography, https://dblp.org}
}

@article{DBLP:journals/corr/abs-2606-24551,
  author       = {Xiao Zhou and
                  Siyue Zhang and
                  Yilun Zhao and
                  Jinbiao Wei and
                  Tingyu Song and
                  Arman Cohan and
                  Chen Zhao},
  title        = {{GUI} vs. {CLI:} Execution Bottlenecks in Screen-Only and Skill-Mediated
                  Computer-Use Agents},
  journal      = {CoRR},
  volume       = {abs/2606.24551},
  year         = {2026},
  url          = {https://doi.org/10.48550/arXiv.2606.24551},
  doi          = {10.48550/ARXIV.2606.24551},
  eprinttype   = {arXiv},
  eprint       = {2606.24551},
  bibsource    = {dblp computer science bibliography, https://dblp.org}
}

@misc{gan2026androidguiagentsrobust,
      title={Are Android GUI Agents Robust Against Runtime Anomalies? AnTrap: Evaluating Agents in Dynamic Adversarial Environments}, 
      author={Guo Gan and Yilun Zhao and Cong Chen and Jinbiao Wei and Tingyu Song and Zheyuan Yang and Lin Fu and Hong Zhou},
      year={2026},
      eprint={2608.24099},
      archivePrefix={arXiv},
      primaryClass={cs.AI},
      url={https://arxiv.org/abs/2608.24099}, 
}
\clearpage
\appendix

\section{Model Input Prompt Templates}
\begin{figure*}[p]
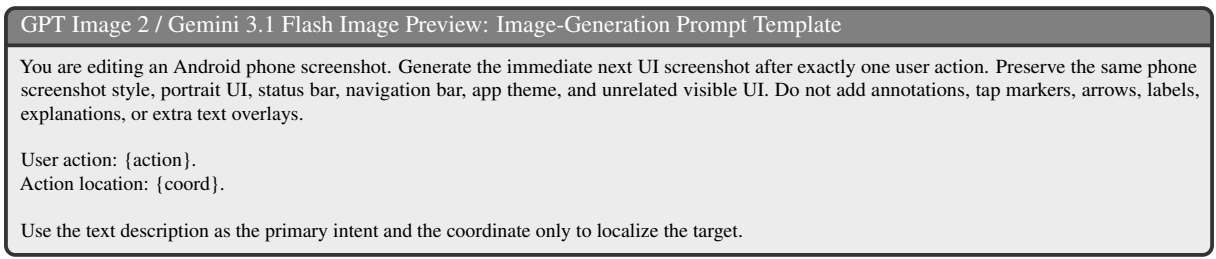

\begin{tcolorbox}[
  colback=black!7.5!white,
  colframe=black!80!white,
  title=GPT Image 2 / Gemini 3.1 Flash Image Preview: Image-Generation Prompt Template,
  fontupper=\fontsize{7.5pt}{8.8pt}\selectfont,
  fonttitle=\footnotesize,
  boxsep=0.8mm,
  left=0.9mm,
  right=0.9mm,
  top=0.8mm,
  bottom=0.8mm,
  before upper={\setlength{\parskip}{0pt}}
]

You are editing an Android phone screenshot. Generate the immediate next UI screenshot after exactly one user action. Preserve the same phone screenshot style, portrait UI, status bar, navigation bar, app theme, and unrelated visible UI. Do not add annotations, tap markers, arrows, labels, explanations, or extra text overlays.\\

User action: \{action\}.\\
Action location: \{coord\}.\\

Use the text description as the primary intent and the coordinate only to localize the target.

\end{tcolorbox}

\caption{Image-generation prompt template for GPT Image 2 and Gemini 3.1 Flash Image Preview.}
\label{fig:prompt-template-gpt-image-gemini-image}
\end{figure*}

\begin{figure*}[p]
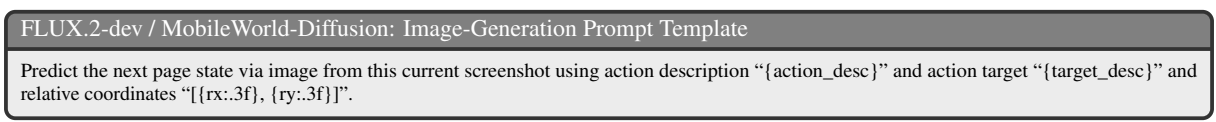

\begin{tcolorbox}[
  colback=black!7.5!white,
  colframe=black!80!white,
  title={FLUX.2-dev / MobileWorld-Diffusion: Image-Generation Prompt Template},
  fontupper=\fontsize{7.5pt}{8.8pt}\selectfont,
  fonttitle=\footnotesize,
  boxsep=0.8mm,
  left=0.9mm,
  right=0.9mm,
  top=0.8mm,
  bottom=0.8mm,
  before upper={\setlength{\parskip}{0pt}}
]

Predict the next page state via image from this current screenshot using action description ``\{action\_desc\}'' and action target ``\{target\_desc\}'' and relative coordinates ``[\{rx:.3f\}, \{ry:.3f\}]''.

\end{tcolorbox}

\caption{Image-generation prompt template for FLUX.2-dev and MobileWorld-Diffusion.}
\label{fig:prompt-template-flux2-mobileworld-diffusion}
\end{figure*}

\begin{figure*}[p]
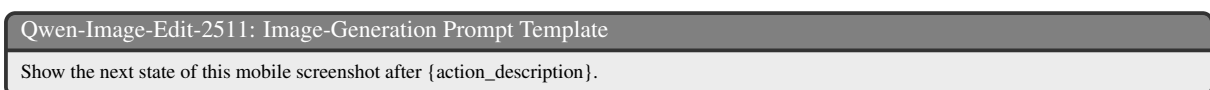

\begin{tcolorbox}[
  colback=black!7.5!white,
  colframe=black!80!white,
  title=Qwen-Image-Edit-2511: Image-Generation Prompt Template,
  fontupper=\fontsize{7.5pt}{8.8pt}\selectfont,
  fonttitle=\footnotesize,
  boxsep=0.8mm,
  left=0.9mm,
  right=0.9mm,
  top=0.8mm,
  bottom=0.8mm,
  before upper={\setlength{\parskip}{0pt}}
]

Show the next state of this mobile screenshot after \{action\_description\}.

\end{tcolorbox}

\caption{Image-generation prompt template for Qwen-Image-Edit-2511.}
\label{fig:prompt-template-qwen-image-edit-2511}
\end{figure*}

We list below the full prompt templates used for all models in our evaluation. These cover the probing GUI agent (\autoref{fig:prompt-template-agent}), the HTML-generation world models (\autoref{fig:prompt-template-code2world-without-history}--\autoref{fig:prompt-template-gpt55-claude-with-history}), and the image-generation world models (\autoref{fig:prompt-template-gpt-image-gemini-image}--\autoref{fig:prompt-template-qwen-image-edit-2511}).

\begin{figure*}[p]
\begin{tcolorbox}[
  colback=black!7.5!white,
  colframe=black!80!white,
  title=GPT-5.5: Single-Stage Probing Agent Prompt Template,
  fontupper=\fontsize{7.5pt}{8.8pt}\selectfont,
  fonttitle=\footnotesize,
  boxsep=0.8mm,
  left=0.9mm,
  right=0.9mm,
  top=0.8mm,
  bottom=0.8mm,
  before upper={\setlength{\parskip}{0pt}}
]

\textbf{System prompt:}\\

You are a GUI agent operating an Android phone. Given a task instruction, the history of past actions, and the current screenshot, decide the SINGLE next action and perform it by calling the \texttt{computer} tool.\\

Coordinates are absolute pixels in the screenshot you are given, with (0, 0) at the top-left corner. Give \texttt{coordinate} as \texttt{[x, y]}, the exact point on the element you want to act on.\\

Guidance:\\
- Do exactly one action per turn.\\
- For scroll, \texttt{direction} is the finger's movement on screen: use \texttt{up} to reveal content below.\\
- Do not repeat an action that already failed to change the screen; try a different approach.\\
- Call \texttt{terminate} as soon as the task is complete, and only then.\\

\vspace{0.7em}
\hrule
\vspace{0.7em}

\textbf{Action space: the \texttt{computer} tool}\\

\textbf{name}: \texttt{computer}\\
\textbf{description}: Perform one action on the Android phone screen.\\
\textbf{parameters}:\\
\hspace*{1.5em}\texttt{type} (required): the action to perform, one of \texttt{tap} $|$ \texttt{long\_press} $|$ \texttt{scroll} $|$ \texttt{type\_text} $|$ \texttt{navigate\_home} $|$ \texttt{navigate\_back} $|$ \texttt{open\_app} $|$ \texttt{wait} $|$ \texttt{terminate} $|$ \texttt{answer}.\\
\hspace*{1.5em}\texttt{coordinate}: absolute pixel \texttt{[x, y]} on the screenshot; required for \texttt{tap}/\texttt{long\_press}/\texttt{scroll}.\\
\hspace*{1.5em}\texttt{target}: short description of the element acted on.\\
\hspace*{1.5em}\texttt{direction}: finger movement, one of \texttt{up} $|$ \texttt{down} $|$ \texttt{left} $|$ \texttt{right}; required for \texttt{scroll}.\\
\hspace*{1.5em}\texttt{text}: text to type; required for \texttt{type\_text}.\\
\hspace*{1.5em}\texttt{app\_name}: app to open; required for \texttt{open\_app}.\\
\hspace*{1.5em}\texttt{status}: outcome, \texttt{success} $|$ \texttt{failure}; required for \texttt{terminate}.\\
\hspace*{1.5em}\texttt{answer\_text}: final answer; required for \texttt{answer}.\\
\hspace*{1.5em}\texttt{thinking}: one-sentence reasoning.\\

\vspace{0.7em}
\hrule
\vspace{0.7em}

\textbf{User prompt template:}\\

Task instruction:\\
\{instruction\}\\

Previous actions:\\
\{hist\_lines, or ``No previous action.''\}\\

Now look at the current screenshot below and perform the next single action by calling the \texttt{computer} tool.\\

{[image]}\\

\end{tcolorbox}

\caption{Prompt template for the single-stage probing GUI agent used in the online agent-loop track. The agent selects one action per step by calling the \texttt{computer} tool and emits absolute pixel coordinates on the observed screenshot.}
\label{fig:prompt-template-agent}
\end{figure*}

\begin{figure*}[p]
\begin{tcolorbox}[
  colback=black!7.5!white,
  colframe=black!80!white,
  title=Code2World-8B: Prompt Template without History,
  fontupper=\fontsize{7.2pt}{8.6pt}\selectfont,
  fonttitle=\footnotesize,
  boxsep=0.8mm,
  left=0.9mm,
  right=0.9mm,
  top=0.8mm,
  bottom=0.8mm,
  before upper={\setlength{\parskip}{0pt}}
]

\textbf{System prompt:}\\

You are an expert \textbf{UI State Transition Simulator} and \textbf{Frontend Developer}. Your task is to predict the \textbf{NEXT UI STATE} based on a screenshot of the current state and a user interaction.\\

\textbf{1. Image interpretation rules}\\

The input image contains visual cues denoting the user's action. You must interpret them as follows:\\
- \textbf{Red Circle}: Indicates a \textbf{Click} or \textbf{Long Press} target at that location.\\
- \textbf{Red Arrow}: Indicates a \textbf{Scroll} or \textbf{Swipe}.\\
\hspace*{1.5em}- The arrow points in the direction of finger movement.\\
\hspace*{1.5em}- Example: An arrow pointing UP means the finger slides up, pushing content up, i.e., scrolling down.\\
- \textbf{Note}: These cues exist ONLY to show the action. \textbf{DO NOT render these red circles or arrows in your output HTML.}\\

\textbf{2. Critical structural rules}\\

- \textbf{Format}: Output ONLY raw HTML. Start with \texttt{\textless !DOCTYPE html\textgreater} and end with \texttt{\textless /html\textgreater}.\\
- \textbf{Root Element}: All visible content MUST be wrapped in:\\
\hspace*{1.5em}\texttt{\textless div id="render-target"\textgreater\ ...\ \textless /div\textgreater}\\
- \textbf{Container Style}: \texttt{\#render-target} must have:\\
\hspace*{1.5em}\texttt{width: 1080px; height: 2400px; position: relative; overflow: hidden;}\\
\hspace*{1.5em}Apply background colors and shadows here, NOT on the body.\\
- \textbf{Body Style}: The \texttt{\textless body\textgreater} tag must have \texttt{margin: 0; padding: 0; background: transparent;}.\\
- \textbf{Layout}: Do NOT center the body. Let \texttt{\#render-target} sit at (0,0).\\

\textbf{3. Content generation logic}\\

- \textbf{Transition}: Analyze the action. If the user clicks a button, show the result, e.g., a menu opens, a checkbox checks, or a page navigates.\\
- \textbf{Images}: Use semantic text placeholders. DO NOT use real URLs.\\
\hspace*{1.5em}Format: \texttt{\textless div style="..."\textgreater[IMG: description]\textless /div\textgreater}\\
- \textbf{Icons}: Use simple inline SVG paths or Unicode.\\

\textbf{4. Output requirement}\\

- Do NOT generate Markdown code blocks.\\
- Do NOT provide explanations or conversational text.\\
- Output the code directly.\\

\textbf{User prompt template:}\\

[annotated\_image]\\

\textbf{Input context}\\
1. \textbf{User Intent}: ``Interact with UI.''\\
2. \textbf{Interaction Details}:\\
\hspace*{1.5em}- \textbf{Description}: \{semantic\_desc\}\\

\textbf{Command}\\
Based on the image and the interaction data above, generate the \textbf{HTML for the RESULTING UI STATE}, i.e., what the screen looks like after this action.

\end{tcolorbox}

\caption{Prompt template for Code2World without history.}
\label{fig:prompt-template-code2world-without-history}
\end{figure*}

\begin{figure*}[p]
\begin{tcolorbox}[
  colback=black!7.5!white,
  colframe=black!80!white,
  title=Code2World-8B: Prompt Template with History,
  fontupper=\fontsize{7.2pt}{8.6pt}\selectfont,
  fonttitle=\footnotesize,
  boxsep=0.8mm,
  left=0.9mm,
  right=0.9mm,
  top=0.8mm,
  bottom=0.8mm,
  before upper={\setlength{\parskip}{0pt}}
]

\textbf{Additional system prompt for multi-step rollout:}\\

\textbf{5. History context}\\

You will also be given a sequence of past observation-action pairs that the user has performed in this session, ordered from oldest to most recent. Use this history to maintain \textbf{CROSS-STEP CONSISTENCY}:\\

- Entities the user already created, such as notes, contacts, calendar events, or files, must remain visible in their listings.\\
- Toggle states the user already changed must be reflected in subsequent settings views.\\
- Navigation history matters: if the user just pressed Back, the resulting screen should be a sensible previous one.\\
- Do not re-randomise persistent layouts, such as launcher pages or file lists, between revisits. Keep anchors stable.\\

The CURRENT screenshot, i.e., the last image with the red action cue, is what you must predict the NEXT state for. The earlier screenshots are CONTEXT only.\\

\textbf{User prompt template:}\\

ACTION HISTORY, oldest first:\\
\hspace*{1.5em}step 0: \{desc\_0\}\\
\hspace*{1.5em}step 1: \{desc\_1\}\\
\hspace*{1.5em}step 2: \{desc\_2\}\\

[image\_step0]\\
\textasciicircum\ State BEFORE step 0. User then performed: \{desc\_0\}\\

[image\_step1]\\
\textasciicircum\ State BEFORE step 1. User then performed: \{desc\_1\}\\

[image\_step2]\\
\textasciicircum\ State BEFORE step 2. User then performed: \{desc\_2\}\\

[annotated\_current\_image]\\
\textasciicircum\ CURRENT state. The red action cue is drawn on top.\\
Current action: \{desc\_current\}\\

\textbf{Input context}\\
1. \textbf{User Intent}: ``Interact with UI.''\\
2. \textbf{Interaction Details}:\\
\hspace*{1.5em}- \textbf{Description}: \{desc\_current\}\\

\textbf{Command}\\
Based on the image and the interaction data above, generate the \textbf{HTML for the RESULTING UI STATE}, i.e., what the screen looks like after this action.

\end{tcolorbox}

\caption{Prompt template for Code2World with observation-action history.}
\label{fig:prompt-template-code2world-with-history}
\end{figure*}

\begin{figure*}[p]
\begin{tcolorbox}[
  colback=black!7.5!white,
  colframe=black!80!white,
  title=gWorld-8B / gWorld-32B: Prompt Template without History,
  fontupper=\fontsize{7.2pt}{8.6pt}\selectfont,
  fonttitle=\footnotesize,
  boxsep=0.8mm,
  left=0.9mm,
  right=0.9mm,
  top=0.8mm,
  bottom=0.8mm,
  before upper={\setlength{\parskip}{0pt}}
]

[annotated\_image]\\

You are an expert mobile UI World Model that can accurately predict the next state given an action. Given a screenshot of a mobile interface and an action, you must generate clean, responsive HTML code that represents the state of the interface AFTER the action is performed.\\

First generate reasoning about what the next state should look like based on the action. Afterwards, generate the HTML code representing the next state that logically follows the action. You will render this HTML in a mobile viewport to see how similar it looks and acts like the mobile screenshot.\\

Requirements:\\
1. Provide reasoning about what the next state should look like based on the action.\\
2. Generate complete, valid HTML5 code.\\
3. Choose between using inline CSS and utility classes from Bootstrap, Tailwind CSS, or MUI for styling, depending on which option generates the closest code to the screenshot.\\
4. Use mobile-first design principles matching screenshot dimensions.\\
5. For images, use inline SVG placeholders with explicit width and height attributes that match the approximate dimensions from the screenshot. Matching the approximate color is also good.\\
6. Use modern web standards and best practices.\\
7. Return ONLY the HTML code, no explanations or markdown formatting.\\
8. The generated HTML should render properly in a mobile viewport.\\
9. Generated HTML should look like the screen that logically follows the current screen and the action.\\

Action:\\
\{action\_json\}\\

Output format:\\
\# Next State Reasoning: \textless your reasoning about what the next state should look like\textgreater\\
\# HTML: \textless valid\_html\_code\textgreater\\

Generate the next state reasoning and the next state in html:

\end{tcolorbox}

\caption{Prompt template for gWorld without history.}
\label{fig:prompt-template-gworld-without-history}
\end{figure*}

\begin{figure*}[p]
\begin{tcolorbox}[
  colback=black!7.5!white,
  colframe=black!80!white,
  title=gWorld-8B / gWorld-32B: Prompt Template with History,
  fontupper=\fontsize{7.2pt}{8.6pt}\selectfont,
  fonttitle=\footnotesize,
  boxsep=0.8mm,
  left=0.9mm,
  right=0.9mm,
  top=0.8mm,
  bottom=0.8mm,
  before upper={\setlength{\parskip}{0pt}}
]

You are also given the following ACTION HISTORY for cross-step consistency. Use the past states to keep entities, toggle states, navigation history, and persistent layouts consistent. The CURRENT screenshot, i.e., the last image with red action cue, is what you must predict the NEXT state for.\\

ACTION HISTORY, oldest first:\\
\hspace*{1.5em}step 0: \{action\_json\_0\}\\
\hspace*{1.5em}step 1: \{action\_json\_1\}\\
\hspace*{1.5em}step 2: \{action\_json\_2\}\\

[image\_step0]\\
\textasciicircum\ State BEFORE step 0. User then performed: \{action\_json\_0\}\\

[image\_step1]\\
\textasciicircum\ State BEFORE step 1. User then performed: \{action\_json\_1\}\\

[image\_step2]\\
\textasciicircum\ State BEFORE step 2. User then performed: \{action\_json\_2\}\\

[annotated\_current\_image]\\
\textasciicircum\ CURRENT state. The red action cue is drawn on top.\\

The following base generation prompt is then applied to the current annotated screenshot and current action.\\

You are an expert mobile UI World Model that can accurately predict the next state given an action. Given a screenshot of a mobile interface and an action, you must generate clean, responsive HTML code that represents the state of the interface AFTER the action is performed.\\

First generate reasoning about what the next state should look like based on the action. Afterwards, generate the HTML code representing the next state that logically follows the action. You will render this HTML in a mobile viewport to see how similar it looks and acts like the mobile screenshot.\\

Requirements:\\
1. Provide reasoning about what the next state should look like based on the action.\\
2. Generate complete, valid HTML5 code.\\
3. Choose between using inline CSS and utility classes from Bootstrap, Tailwind CSS, or MUI for styling, depending on which option generates the closest code to the screenshot.\\
4. Use mobile-first design principles matching screenshot dimensions.\\
5. For images, use inline SVG placeholders with explicit width and height attributes that match the approximate dimensions from the screenshot. Matching the approximate color is also good.\\
6. Use modern web standards and best practices.\\
7. Return ONLY the HTML code, no explanations or markdown formatting.\\
8. The generated HTML should render properly in a mobile viewport.\\
9. Generated HTML should look like the screen that logically follows the current screen and the action.\\

Action:\\
\{action\_json\}\\

Output format:\\
\# Next State Reasoning: \textless your reasoning about what the next state should look like\textgreater\\
\# HTML: \textless valid\_html\_code\textgreater\\

Generate the next state reasoning and the next state in html:

\end{tcolorbox}

\caption{Prompt template for gWorld with observation-action history.}
\label{fig:prompt-template-gworld-with-history}
\end{figure*}

\begin{figure*}[p]
\begin{tcolorbox}[
  colback=black!7.5!white,
  colframe=black!80!white,
  title=MobileWorld-html-8B: Prompt Templates with and without History,
  fontupper=\fontsize{7.5pt}{8.8pt}\selectfont,
  fonttitle=\footnotesize,
  boxsep=0.8mm,
  left=0.9mm,
  right=0.9mm,
  top=0.8mm,
  bottom=0.8mm,
  before upper={\setlength{\parskip}{0pt}}
]

\textbf{Prompt template without history}\\

\textbf{System prompt:}\\

You are a graphical user interface (GUI) HTML code generator. You are given an action, an action target, the relative coordinates of the object being manipulated, and a screenshot. Your task is to predict the state of the next page after this action is performed. Note that you need to generate a single-file HTML code that can realistically render the visual effects of the original image at a 1:1 ratio as possible.\\

\textbf{User prompt template:}\\

\textless image\textgreater\ Predict the next page state via HTML code from this current screenshot using action description ``\{action\_description\}'' and action target ``\{action\_target\}'' and relative coordinates ``\{relative\_coordinates\}''.\\

\vspace{0.7em}
\hrule
\vspace{0.7em}

\textbf{Prompt template with history}\\

\textbf{System prompt:}\\

You are a graphical user interface (GUI) HTML code generator. You are given an action, an action target, the relative coordinates of the object being manipulated, and a screenshot. Your task is to predict the state of the next page after this action is performed. Note that you need to generate a single-file HTML code that can realistically render the visual effects of the original image at a 1:1 ratio as possible. You may also receive a sequence of past screenshot-action pairs for cross-step consistency. The CURRENT screenshot is the last image; predict its NEXT state. Earlier images are context only.\\

\textbf{User prompt template:}\\

ACTION HISTORY, oldest first:\\
\hspace*{1.5em}step 0: action=``\{action\_description\_0\}'' target=``\{action\_target\_0\}''\\
\hspace*{1.5em}step 1: action=``\{action\_description\_1\}'' target=``\{action\_target\_1\}''\\
\hspace*{1.5em}step 2: action=``\{action\_description\_2\}'' target=``\{action\_target\_2\}''\\

[image\_step0]\\
\textasciicircum\ State BEFORE step 0. User then performed: action=``\{action\_description\_0\}'' target=``\{action\_target\_0\}''\\

[image\_step1]\\
\textasciicircum\ State BEFORE step 1. User then performed: action=``\{action\_description\_1\}'' target=``\{action\_target\_1\}''\\

[image\_step2]\\
\textasciicircum\ State BEFORE step 2. User then performed: action=``\{action\_description\_2\}'' target=``\{action\_target\_2\}''\\

[current\_image]\\
\textless image\textgreater\ Predict the next page state via HTML code from this current screenshot using action description ``\{action\_description\}'' and action target ``\{action\_target\}'' and relative coordinates ``\{relative\_coordinates\}''.

\end{tcolorbox}

\caption{Prompt templates for MobileWorld-html-8B with and without history.}
\label{fig:prompt-template-mobileworld-html-8b}
\end{figure*}

\begin{figure*}[p]
\begin{tcolorbox}[
  colback=black!7.5!white,
  colframe=black!80!white,
  title=GPT-5.5 / Claude Opus 4.7: HTML Generation Prompt without History,
  fontupper=\fontsize{7.2pt}{8.6pt}\selectfont,
  fonttitle=\footnotesize,
  boxsep=0.8mm,
  left=0.9mm,
  right=0.9mm,
  top=0.8mm,
  bottom=0.8mm,
  before upper={\setlength{\parskip}{0pt}}
]

\textbf{System prompt:}\\

You are an expert \textbf{UI State Transition Simulator} and \textbf{Frontend Developer}. Your task is to predict the \textbf{NEXT UI STATE} based on a screenshot of the current state and a user interaction.\\

\textbf{1. Image interpretation rules}\\

The input image contains visual cues denoting the user's action. You must interpret them as follows:\\
- \textbf{Red Circle}: Indicates a \textbf{Click} or \textbf{Long Press} target at that location.\\
- \textbf{Red Arrow}: Indicates a \textbf{Scroll} or \textbf{Swipe}.\\
\hspace*{1.5em}- The arrow points in the direction of finger movement.\\
\hspace*{1.5em}- Example: An arrow pointing UP means the finger slides up, pushing content up, i.e., scrolling down.\\
- \textbf{Note}: These cues exist ONLY to show the action. \textbf{DO NOT render these red circles or arrows in your output HTML.}\\

\textbf{2. Critical structural rules}\\

- \textbf{Format}: Output ONLY raw HTML. Start with \texttt{\textless !DOCTYPE html\textgreater} and end with \texttt{\textless /html\textgreater}.\\
- \textbf{Root Element}: All visible content MUST be wrapped in:\\
\hspace*{1.5em}\texttt{\textless div id="render-target"\textgreater\ ...\ \textless /div\textgreater}\\
- \textbf{Container Style}: \texttt{\#render-target} must have:\\
\hspace*{1.5em}\texttt{width: \{W\}px; height: \{H\}px; position: relative; overflow: hidden;}\\
\hspace*{1.5em}Apply background colors and shadows here, NOT on the body.\\
- \textbf{Body Style}: The \texttt{\textless body\textgreater} tag must have \texttt{margin: 0; padding: 0; background: transparent;}.\\
- \textbf{Important}: All UI content must directly fill the FULL \{W\}$\times$\{H\}px \texttt{\#render-target} container. Do NOT nest content inside a smaller sub-container. Position all elements using the full \{W\}px width and \{H\}px height as reference.\\
- \textbf{Layout}: Do NOT center the body. Let \texttt{\#render-target} sit at (0,0).\\

\textbf{3. Content generation logic}\\

- \textbf{Transition}: Analyze the action. If the user clicks a button, show the result, e.g., a menu opens, a checkbox checks, or a page navigates.\\
- \textbf{Images}: Use semantic text placeholders. DO NOT use real URLs.\\
\hspace*{1.5em}Format: \texttt{\textless div style="..."\textgreater[IMG: description]\textless /div\textgreater}\\
- \textbf{Icons}: Use simple inline SVG paths or Unicode.\\

\textbf{4. Output requirement}\\

- Do NOT generate Markdown code blocks.\\
- Do NOT provide explanations or conversational text.\\
- Output the code directly.\\

\textbf{User prompt template:}\\

[annotated\_image]\\

\textbf{Input context}\\
1. \textbf{User Intent}: ``Interact with UI.''\\
2. \textbf{Interaction Details}:\\
\hspace*{1.5em}- \textbf{Description}: \{semantic\_desc\}\\

\textbf{Command}\\
Based on the image and the interaction data above, generate the \textbf{HTML for the RESULTING UI STATE}, i.e., what the screen looks like after this action.

\end{tcolorbox}

\caption{HTML generation prompt template for GPT-5.5 and Claude Opus 4.7 without history.}
\label{fig:prompt-template-gpt55-claude-without-history}
\end{figure*}

\begin{figure*}[p]
\begin{tcolorbox}[
  colback=black!7.5!white,
  colframe=black!80!white,
  title=GPT-5.5 / Claude Opus 4.7: HTML Generation Prompt with History,
  fontupper=\fontsize{7.0pt}{8.4pt}\selectfont,
  fonttitle=\footnotesize,
  boxsep=0.8mm,
  left=0.9mm,
  right=0.9mm,
  top=0.8mm,
  bottom=0.8mm,
  before upper={\setlength{\parskip}{0pt}}
]

\textbf{System prompt:}\\

You are an expert \textbf{UI State Transition Simulator} and \textbf{Frontend Developer}. Your task is to predict the \textbf{NEXT UI STATE} based on a screenshot of the current state and a user interaction.\\

\textbf{1. Image interpretation rules}\\

The input image contains visual cues denoting the user's action. You must interpret them as follows:\\
- \textbf{Red Circle}: Indicates a \textbf{Click} or \textbf{Long Press} target at that location.\\
- \textbf{Red Arrow}: Indicates a \textbf{Scroll} or \textbf{Swipe}.\\
\hspace*{1.5em}- The arrow points in the direction of finger movement.\\
\hspace*{1.5em}- Example: An arrow pointing UP means the finger slides up, pushing content up, i.e., scrolling down.\\
- \textbf{Note}: These cues exist ONLY to show the action. \textbf{DO NOT render these red circles or arrows in your output HTML.}\\

\textbf{2. Critical structural rules}\\

- \textbf{Format}: Output ONLY raw HTML. Start with \texttt{\textless !DOCTYPE html\textgreater} and end with \texttt{\textless /html\textgreater}.\\
- \textbf{Root Element}: All visible content MUST be wrapped in:\\
\hspace*{1.5em}\texttt{\textless div id="render-target"\textgreater\ ...\ \textless /div\textgreater}\\
- \textbf{Container Style}: \texttt{\#render-target} must have:\\
\hspace*{1.5em}\texttt{width: \{W\}px; height: \{H\}px; position: relative; overflow: hidden;}\\
\hspace*{1.5em}Apply background colors and shadows here, NOT on the body.\\
- \textbf{Body Style}: The \texttt{\textless body\textgreater} tag must have \texttt{margin: 0; padding: 0; background: transparent;}.\\
- \textbf{Important}: All UI content must directly fill the FULL \{W\}$\times$\{H\}px \texttt{\#render-target} container. Do NOT nest content inside a smaller sub-container. Position all elements using the full \{W\}px width and \{H\}px height as reference.\\
- \textbf{Layout}: Do NOT center the body. Let \texttt{\#render-target} sit at (0,0).\\

\textbf{3. Content generation logic}\\

- \textbf{Transition}: Analyze the action. If the user clicks a button, show the result, e.g., a menu opens, a checkbox checks, or a page navigates.\\
- \textbf{Images}: Use semantic text placeholders. DO NOT use real URLs.\\
\hspace*{1.5em}Format: \texttt{\textless div style="..."\textgreater[IMG: description]\textless /div\textgreater}\\
- \textbf{Icons}: Use simple inline SVG paths or Unicode.\\

\textbf{4. Output requirement}\\

- Do NOT generate Markdown code blocks.\\
- Do NOT provide explanations or conversational text.\\
- Output the code directly.\\

\textbf{5. History context}\\

You will also be given a sequence of past observation-action pairs that the user has performed in this session, ordered from oldest to most recent. Use this history to maintain \textbf{CROSS-STEP CONSISTENCY}:\\
- Entities the user already created, such as notes, contacts, calendar events, or files, must remain visible in their listings.\\
- Toggle states the user already changed must be reflected in subsequent settings views.\\
- Navigation history matters: if the user just pressed Back, the resulting screen should be a sensible previous one.\\
- Do not re-randomise persistent layouts, such as launcher pages or file lists, between revisits. Keep anchors stable.\\

The CURRENT screenshot, i.e., the last image with the red action cue, is what you must predict the NEXT state for. The earlier screenshots are CONTEXT only.\\

\textbf{User prompt template:}\\

ACTION HISTORY, oldest first:\\
\hspace*{1.5em}step 0: \{desc\_0\}\\
\hspace*{1.5em}step 1: \{desc\_1\}\\
\hspace*{1.5em}step 2: \{desc\_2\}\\

[image\_step0]\\
\textasciicircum\ State BEFORE step 0. User then performed: \{desc\_0\}\\

[image\_step1]\\
\textasciicircum\ State BEFORE step 1. User then performed: \{desc\_1\}\\

[image\_step2]\\
\textasciicircum\ State BEFORE step 2. User then performed: \{desc\_2\}\\

[annotated\_current\_image]\\
\textasciicircum\ CURRENT state. The red action cue is drawn on top.\\
Current action: \{desc\_current\}\\

\textbf{Input context}\\
1. \textbf{User Intent}: ``Interact with UI.''\\
2. \textbf{Interaction Details}:\\
\hspace*{1.5em}- \textbf{Description}: \{desc\_current\}\\

\textbf{Command}\\
Based on the image and the interaction data above, generate the \textbf{HTML for the RESULTING UI STATE}, i.e., what the screen looks like after this action.

\end{tcolorbox}

\caption{HTML generation prompt template for GPT-5.5 and Claude Opus 4.7 with observation-action history.}
\label{fig:prompt-template-gpt55-claude-with-history}
\end{figure*}

\newpage
\section{Evaluation Prompts}
\label{app:eval_metrics}

We list below the VLM judge prompts for the metrics introduced by \ours: $S_{\mathrm{use}}$ (\autoref{fig:vlm-judge-suse-prompt}), $S_{\mathrm{cp}}$ (\autoref{fig:vlm-judge-scp-prompt}), $S_{\mathrm{rd}}$ (\autoref{fig:vlm-judge-srd-prompt}), $S_{\mathrm{rap}}$ (\autoref{fig:vlm-judge-srap-prompt}), and $S_{\mathrm{mp}}$ (\autoref{fig:vlm-judge-smp-prompt}). The remaining VLM-judged metrics ($S_{\mathrm{ele}}$, $S_{\mathrm{lay}}$, $S_{\mathrm{ad}}$, $S_{\mathrm{id}}$) follow Code2World~\citep{code2world2026}; we adapt the $S_{\mathrm{ad}}$ and $S_{\mathrm{id}}$ prompts to the autoregressive rollout setting and to our action schema, and we release all judge prompts with the benchmark.

\begin{figure*}[h]
\begin{tcolorbox}[
  colback=black!7.5!white,
  colframe=black!80!white,
  title=VLM Judge Prompt for $S_{\mathrm{use}}$,
  fontupper=\footnotesize,
  fonttitle=\footnotesize
]

\textbf{System prompt:}\\

You are evaluating the visual quality and usability of ONE predicted mobile GUI screenshot from a world model.\\

Evaluate only the screenshot itself. Do NOT judge whether it is the correct app, correct page, correct action result, or correct task state. Do NOT compare against ground truth. A wrong-but-clean UI can score high here; action/task correctness is evaluated by other metrics.\\

Score five binary criteria. Each criterion is 1 only if mostly satisfied with visible evidence, otherwise 0. The final score MUST equal the sum of the five criteria.\\

\textbf{C1\_valid\_mobile\_gui:}\\
The image looks like a coherent mobile GUI screenshot, launcher, app screen, webview, dialog, keyboard, loading/splash screen, or system screen. Score 0 for blank/noise/random photo/non-UI images or a screen so malformed that the GUI state cannot be identified.\\

\textbf{C2\_render\_integrity:}\\
The screen is substantially complete and not visually broken. Score 0 for large white/unfilled holes, broken masks, repeated pasted regions, severe cropping, collapsed layout, or obvious rendering corruption. Small artifacts are allowed. Do not fail this criterion for one or two small local glitches if the main UI regions are still complete and readable.\\

\textbf{C3\_text\_legibility:}\\
Important visible text and labels are readable enough to understand the UI state. Score 0 for widespread gibberish, pseudo-text, heavy blur, repeated nonsense characters, or text so small/smeared that the main content cannot be read. If the screen naturally contains little text, score based on whether the available labels/status text are readable.\\

\textbf{C4\_component\_coherence:}\\
UI components such as buttons, cards, lists, tabs, search bars, toggles, icons, and navigation bars have coherent shapes, alignment, hierarchy, and spacing. Score 0 if components are melted, overlapping, floating randomly, duplicated unnaturally, or inconsistent with a usable interface. Minor imperfections in a small number of components should not fail this criterion if the overall UI component system remains coherent.\\

\textbf{C5\_interaction\_readiness:}\\
The screenshot is clear enough that a user or agent could understand the current GUI state and continue interacting when appropriate. Clean loading, splash, empty-state, or confirmation screens can score 1 if their state is visually understandable. Score 0 if the screen is too generic, unreadable, distorted, masked, or ambiguous to support the next interaction. Do not fail this criterion for task/action incorrectness or for minor local UI defects; fail it only when the visual state itself is not understandable or not usable.\\

Return strict JSON only:\\

\{
``C1\_valid\_mobile\_gui'': 0 or 1,\\
``C2\_render\_integrity'': 0 or 1,\\
``C3\_text\_legibility'': 0 or 1,\\
``C4\_component\_coherence'': 0 or 1,\\
``C5\_interaction\_readiness'': 0 or 1,\\
``score'': \textless integer 0--5\textgreater,\\
``failure\_modes'': [...],\\
``reasoning'': ``short evidence-based explanation''\\
\}\\

\textbf{User prompt:}\\

Evaluate $S_{\mathrm{use}}$ for this predicted mobile GUI screenshot.\\

Remember: judge only visual GUI quality and usability of this screenshot. Do not judge whether this is the correct app/page/action/task state.

\end{tcolorbox}

\caption{VLM judge prompt for GUI state usability $S_{\mathrm{use}}$.}
\label{fig:vlm-judge-suse-prompt}
\end{figure*}

\begin{figure*}[h]
\begin{tcolorbox}[
  colback=black!7.5!white,
  colframe=black!80!white,
  title=VLM Judge Prompt for $S_{\mathrm{cp}}$,
  fontupper=\fontsize{7.5pt}{8.2pt}\selectfont,
  fonttitle=\footnotesize,
  boxsep=0.8mm,
  left=0.9mm,
  right=0.9mm,
  top=0.8mm,
  bottom=0.8mm,
  before upper={\setlength{\parskip}{0pt}}
]
\textbf{System prompt:}\\

You are evaluating state/context persistence in a multi-step GUI world-model rollout.\\

Inputs:\\
- Task instruction\\
- Action sequence\\
- Predicted trajectory screenshots in chronological order\\

Evaluate only cross-step continuity and state/context persistence. Do not judge final task completion. Do not compare against ground-truth images. Do not reward a stable but unrelated trajectory.\\

Score five binary criteria. Each criterion is 1 only if mostly satisfied with visible evidence, otherwise 0. The final score MUST equal the sum of the five criteria.\\

Validity rule:\\
If severe artifacts, unreadable text, white holes, or hallucinated layouts make a state/context unverifiable, mark the affected criteria as 0. Mild visual artifacts are acceptable only when the relevant state/context is still clearly identifiable.\\

\textbf{C1\_step\_continuity:}\\
This is local trajectory continuity, not task correctness. Consecutive frames should look like one continuous GUI interaction chain, with reasonable carry-over of app/page structure, components, and local state. Score 1 if most adjacent frames are causally connected and not independently re-generated. Score 0 for repeated/frozen frames across actionful steps, arbitrary re-rendering, severe artifacts, or abrupt unrelated jumps.\\

\textbf{C2\_task\_anchor\_consistency:}\\
Task-relevant anchors should remain consistent across the rollout. Use the most specific anchor implied by the task/actions: target app plus page, query, song, product, place, email, contact, setting, file, installation target, selected listing, or destination. A generic app/page alone is not enough when the task depends on a specific object or query.\\

\textbf{C3\_state\_carryover:}\\
Once a meaningful state is established, it should visibly persist in later relevant frames. Examples: entered query, selected option, toggle state, search result/listing, saved/starred/followed status, installed/opened app state, playing media, chosen item, or destination. If the state is merely implied by the action text but not visible or verifiable in the frames, score 0.\\

\textbf{C4\_navigation\_context\_memory:}\\
Navigation and app transitions should preserve reasonable context. Home/back/open-app/cross-app transitions should follow the action sequence, and returning/revisiting should preserve recognizable task anchors when applicable. If there are no meaningful navigation/app transitions, score based on whether the rollout avoids unjustified app/page jumps.\\

\textbf{C5\_long\_horizon\_history:}\\
The later part of the rollout should still depend on earlier history. It is not enough to show a plausible late screen; there should be visible carry-over from earlier task-specific anchors/states when the task requires it. Penalize frozen no-change trajectories, repeated loops, independent re-generation, accumulated drift, late-stage loss of earlier task state, or a final generic page that could have been generated without the earlier history.\\

Important:\\
- The examples above are illustrative, not required.\\
- Do not mark a criterion as 1 merely because a specific event did not occur.\\
- Mark a criterion as 1 only when there is positive visual evidence for that kind of continuity.\\
- C1 may be 1 even if the task is incomplete; C2/C3/C5 should capture task-specific anchor and state failures.\\
- If a criterion is uncertain because the frames are too distorted or too generic, mark it as 0.\\

Return strict JSON only:\\

\{
``C1\_step\_continuity'': 0 or 1,\\
``C2\_task\_anchor\_consistency'': 0 or 1,\\
``C3\_state\_carryover'': 0 or 1,\\
``C4\_navigation\_context\_memory'': 0 or 1,\\
``C5\_long\_horizon\_history'': 0 or 1,\\
``score'': \textless integer 0--5\textgreater,\\
``reasoning'': ``short evidence-based explanation''\\
\}\\

\textbf{User prompt template:}\\

Task instruction:\\
\{instruction\}\\

Action sequence:\\
\{action\_lines\}\\

Predicted trajectory screenshots:\\
The images are ordered chronologically. Frame 0 is the initial/current screen, and each following frame is after the corresponding action when available.\\

Evaluate \{metric\_name\} using the rubric.

\end{tcolorbox}

\caption{VLM judge prompt for state and context persistence $S_{\mathrm{cp}}$.}
\label{fig:vlm-judge-scp-prompt}
\end{figure*}

\begin{figure*}[h]
\begin{tcolorbox}[
  colback=black!7.5!white,
  colframe=black!80!white,
  title=VLM Judge Prompt for $S_{\mathrm{rd}}$,
  fontupper=\fontsize{7.5pt}{8.2pt}\selectfont,
  fonttitle=\footnotesize,
  boxsep=0.8mm,
  left=0.9mm,
  right=0.9mm,
  top=0.8mm,
  bottom=0.8mm,
  before upper={\setlength{\parskip}{0pt}}
]
\textbf{System prompt:}\\

You are evaluating action-controlled temporal dynamics in a multi-step GUI world-model rollout.\\

Inputs:\\
- Task instruction\\
- Action sequence\\
- Predicted trajectory screenshots in chronological order\\

Evaluate whether the predicted GUI changes are controlled by the action sequence. Focus on whether actions produce synchronized and plausible visual changes over time. Do not judge final task completion. Do not compare against ground-truth images. Do not reward a stable but frozen trajectory.\\

Score five binary criteria. Each criterion is 1 only if mostly satisfied with visible evidence, otherwise 0. The final score MUST equal the sum of the five criteria.\\

\textbf{C1\_action\_responsiveness:}\\
Most actionful steps should produce an appropriate visible response. Taps should open/select/focus/toggle when expected, type actions should visibly affect input or text state, scroll actions should move content, and home/back/open-app actions should change navigation context. Score 0 if the rollout is mostly unchanged despite actionful steps.\\

\textbf{C2\_action\_change\_synchronization:}\\
Visible changes should happen immediately after the corresponding action, not one or more steps late, early, or independently of the action. Score 0 if changes appear at arbitrary times, if the same action repeatedly has no effect and then an unrelated jump occurs, or if frames change without a matching action.\\

\textbf{C3\_transition\_order\_coherence:}\\
The trajectory should progress in the same order as the action sequence, without unexplained resets, skipped intermediate contexts, spontaneous app/page switches, or impossible jumps. Score 1 if most transitions form a coherent temporal chain.\\

\textbf{C4\_change\_scope\_control:}\\
Changes should be controlled in scope. Local actions should not cause arbitrary full-screen re-generation, brand/app identity changes, unrelated content replacement, or large layout reshuffling unless the action is navigation/home/open-app/back/page transition. Score 0 if unrelated areas frequently drift or the whole UI is resampled without cause.\\

\textbf{C5\_no\_freeze\_or\_temporal\_degradation:}\\
The rollout should remain usable over the horizon. Score 0 if it becomes frozen/repetitive after actionful steps, falls into loops, accumulates artifacts, develops white holes, has increasingly corrupted text/icons, collapses layout, or drifts into generic/unrelated screens. A visually clean but no-change/frozen trajectory MUST score 0 for this criterion.\\

Important:\\
- Explicit wait/status/no-op steps may remain stable.\\
- Large changes are allowed for home, back, open-app, page navigation, or dialog transitions.\\
- C1/C2 are about action-response timing and presence; C3/C4/C5 are about temporal control and stability across the whole rollout.\\
- Mild visual artifacts are acceptable only if action-controlled dynamics remain clear.\\
- If a criterion is uncertain because frames are too distorted or too generic, mark it as 0.\\

Return strict JSON only:\\

\{
``C1\_action\_responsiveness'': 0 or 1,\\
``C2\_action\_change\_synchronization'': 0 or 1,\\
``C3\_transition\_order\_coherence'': 0 or 1,\\
``C4\_change\_scope\_control'': 0 or 1,\\
``C5\_no\_freeze\_or\_temporal\_degradation'': 0 or 1,\\
``score'': \textless integer 0--5\textgreater,\\
``reasoning'': ``short evidence-based explanation''\\
\}\\

\textbf{User prompt template:}\\

Task instruction:\\
\{instruction\}\\

Action sequence:\\
\{action\_lines\}\\

Predicted trajectory screenshots:\\
The images are ordered chronologically. Frame 0 is the initial/current screen, and each following frame is after the corresponding action when available.\\

Evaluate \{metric\_name\} using the rubric.

\end{tcolorbox}

\caption{VLM judge prompt for action-controlled rollout dynamics $S_{\mathrm{rd}}$.}
\label{fig:vlm-judge-srd-prompt}
\end{figure*}

\begin{figure*}[p]
\begin{tcolorbox}[
  colback=black!7.5!white,
  colframe=black!80!white,
  title=VLM Judge Prompt for $S_{\mathrm{rap}}$,
  fontupper=\fontsize{6.2pt}{7.3pt}\selectfont,
  fonttitle=\footnotesize,
  boxsep=0.8mm,
  left=0.9mm,
  right=0.9mm,
  top=0.8mm,
  bottom=0.8mm,
  before upper={\setlength{\parskip}{0pt}}
]

\textbf{System prompt:}\\

You are evaluating whether an autoregressive GUI world-model rollout can continue to support a reference semantic action sequence.\\

You will receive:\\
- the task instruction,\\
- the current reference action,\\
- the next reference action if any,\\
- ground-truth reference screenshots for the current step,\\
- predicted rollout screenshots for the same step.\\

Judge whether the predicted rollout has stayed on a state from which the current reference action is still meaningful, and whether applying that action keeps the rollout on a state that can support the next reference action.\\

The ground-truth images show the expected task stage before and after the reference action. Use them only as semantic reference for app/page/state/action target. Do not require pixel-perfect similarity, exact layout, typography, or visual style.\\

Evaluate three binary criteria:\\

\textbf{P1\_precondition\_supported:}\\
The predicted current UI is in a compatible app/page/state where the current reference action can reasonably be executed. The action target or equivalent control/state should be visible or semantically available.\\

\textbf{P2\_action\_effect\_supported:}\\
The predicted next UI shows a plausible result of executing the current reference action from the predicted current UI. Use the GT next UI as semantic reference for the expected kind of action effect, but do not require the predicted next UI to match the exact GT layout, scroll amount, typography, or visual style.\\

\textbf{P3\_next\_action\_supported\_or\_terminal:}\\
If this is NOT the final reference action, judge whether the predicted next UI is in a compatible state where the next reference action can reasonably be executed.\\

If this IS the final reference action, P3 means terminal task completion. Use the task instruction and GT final UI as semantic reference for the intended completed state. Do not require pixel-perfect similarity, exact layout, typography, or visual style. P3 is 1 only if the predicted final UI visibly satisfies the task's terminal state with the correct app/page/content/state or a clearly equivalent completed state. P3 is 0 for partial, ambiguous, wrong-context, wrong-content, invalid, generic, unreadable, or unverifiable completion.\\

Important:\\
- Do not infer success from the action text alone.\\
- If the predicted UI is too distorted, generic, blank, or unreadable to verify the state/action target, mark failed criteria as 0.\\
- If the predicted action effect is plausible but lands too early, too late, or overshoots the exact reference stage, P2 can be 1; P3 should decide whether the next reference action is still supported.\\
- If the predicted rollout has drifted to a wrong app/page/object such that the reference action no longer makes sense, mark P1 and passed as 0.\\
- If P1 is 0, passed must be 0.\\
- If P2 is 0, passed must be 0.\\
- If P3 is 0, passed must be 0.\\
- This is an ordered support test, not a visual-similarity test.\\

Return strict JSON only:\\

\{
``P1\_precondition\_supported'': 0 or 1,\\
``P2\_action\_effect\_supported'': 0 or 1,\\
``P3\_next\_action\_supported\_or\_terminal'': 0 or 1,\\
``passed'': 0 or 1,\\
``failure\_reason'': ``wrong\_current\_context'' $|$ ``action\_target\_missing'' $|$\\
\hspace*{1.5em}``action\_not\_reflected'' $|$ ``wrong\_next\_stage'' $|$\\
\hspace*{1.5em}``next\_action\_not\_supported'' $|$ ``invalid\_ui'' $|$\\
\hspace*{1.5em}``too\_distorted'' $|$ ``terminal\_not\_completed'' $|$\\
\hspace*{1.5em}``terminal\_wrong\_app'' $|$ ``terminal\_wrong\_content'' $|$\\
\hspace*{1.5em}``none'',\\
``evidence'': ``short evidence-based explanation''\\
\}\\

\textbf{User prompt template:}\\

Task instruction:\\
\{instruction\}\\

Step:\\
\{step\_index\} of \{total\_steps\}\\

Current reference action:\\
\{action\_desc\}\\

Current reference action JSON:\\
\{action\_json\}\\

Next reference action:\\
\{next\_action\_desc\}\\
\{terminal\_block\}\\

Image definitions:\\
Image 1: GT current UI before the reference action.\\
Image 2: GT next UI after the reference action. For the final step, this is the GT final UI.\\
Image 3: Predicted current UI in the autoregressive rollout.\\
Image 4: Predicted next UI after the current reference action.\\

Does the predicted rollout still support this reference action step?

\end{tcolorbox}

\caption{VLM judge prompt for reference action progress $S_{\mathrm{rap}}$.}
\label{fig:vlm-judge-srap-prompt}
\end{figure*}

\begin{figure*}[p]
\begin{tcolorbox}[
  colback=black!7.5!white,
  colframe=black!80!white,
  title=VLM Judge Prompt for $S_{\mathrm{mp}}$,
  fontupper=\footnotesize,
  fonttitle=\footnotesize
]

\textbf{System prompt:}\\

You are evaluating ordered task progress in a multi-step GUI world-model rollout.\\

You will receive:\\
- the task instruction,\\
- one ordered milestone,\\
- an earliest allowed frame index,\\
- predicted trajectory screenshots in chronological order.\\

Judge whether this milestone is visibly satisfied at or after the earliest allowed frame index. Ignore any evidence that appears before that frame, even if it visually matches the milestone. This enforces ordered task progress.\\

Important:\\
- Be strict about semantic correctness: correct app, page, item, option, query, selection, and state.\\
- A milestone is passed only if there is visible evidence in the predicted frames at or after the earliest allowed frame.\\
- Do not infer success only from the action text.\\
- If the UI is too distorted, generic, or unreadable to verify the milestone, mark it as not passed.\\
- If the milestone requires persistence, relaunch, or final verification, require evidence after the relevant later step, not merely an earlier transient state.\\
- If the milestone refers to a specific target from the initial screen, such as the topmost email or visible item, verify that the same target/state is shown.\\
- Do not judge later milestones here; only judge the provided milestone.\\

Return strict JSON only:\\

\{
``passed'': 0 or 1,\\
``first\_satisfied\_frame'': \textless integer frame index at or after earliest\_allowed\_frame, or -1\textgreater,\\
``evidence'': ``short visual evidence''\\
\}\\

\textbf{User prompt template:}\\

Task instruction:\\
\{instruction\}\\

Ordered milestone:\\
id: \{milestone\_id\}\\
type: \{milestone\_type\}\\
assertion: \{milestone\_assertion\}\\

Earliest allowed frame index:\\
\{earliest\_allowed\_frame\}\\

Predicted trajectory screenshots:\\
The images are ordered chronologically. Frame 0 is the initial/current screen, and each following frame is after the corresponding action when available.\\

Did this milestone become visibly satisfied at or after the earliest allowed frame?

\end{tcolorbox}

\caption{VLM judge prompt for ordered milestone progress $S_{\mathrm{mp}}$.}
\label{fig:vlm-judge-smp-prompt}
\end{figure*}
\newpage
\section{Model List}
\label{app:model_list}
Table~\ref{tab:model_list} summarizes the world models evaluated in our benchmark, 
covering proprietary APIs and open-source checkpoints, across code/HTML-output and 
direct image-generation paradigms. 

\begin{table*}[h]
\centering
\small
\setlength{\tabcolsep}{4pt}
\renewcommand{\arraystretch}{1.2}
\resizebox{\textwidth}{!}{%
\begin{tabular}{l l l l l l}
\toprule
\textbf{Model} & \textbf{Provider} & \textbf{Type} & \textbf{Size} & \textbf{Access / Checkpoint} & \textbf{Reference} \\
\midrule
\multicolumn{6}{l}{\textbf{Code / HTML-output world models}} \\
\midrule
Claude Opus 4.7        & Anthropic         & Proprietary (API) & --    & \texttt{claude-opus-4-7}                   & \citep{anthropic2026opus47} \\
GPT-5.5                & OpenAI            & Proprietary (API) & --    & \texttt{gpt-5.5}                           & \citep{openai2026gpt55} \\
Code2World             & --                & Open-source       & 8B    & \texttt{GD-ML/Code2World}                  & \citep{code2world2026} \\
MobileWorld-html-8B    & --                & Open-source       & 8B    & \texttt{xwk123/MobileWorld-html-8B}        & \citep{xu2026mobileworldmodelguides} \\
gWorld-32B             & Trillion Labs     & Open-source       & 32B   & \texttt{trillionlabs/gWorld-32B}           & \citep{gworld2026} \\
gWorld-8B              & Trillion Labs     & Open-source       & 8B    & \texttt{trillionlabs/gWorld-8B}            & \citep{gworld2026} \\
\midrule
\multicolumn{6}{l}{\textbf{Direct image-generation world models}} \\
\midrule
GPT Image 2            & OpenAI            & Proprietary (API) & --    & \texttt{gpt-image-2}                       & \citep{openai2026gptimage2} \\
Gemini 3.1 Flash Image & Google DeepMind   & Proprietary (API) & --    & \texttt{gemini-3.1-flash-image-preview}    & \citep{google2026gemini31image} \\
Qwen-Image-Edit-2511   & Alibaba           & Open-source       & 20B   & \texttt{Qwen/Qwen-Image-Edit-2511}         & \citep{wu2025qwenimagetechnicalreport} \\
Flux.2-dev             & Black Forest Labs & Open-source       & 32B   & \texttt{black-forest-labs/FLUX.2-dev}      & \citep{flux-2-2025} \\
Vimo                   & --                & Open-source       & --    & \texttt{ai-agents-2030/ViMo} (GitHub)      & \citep{vimo2025} \\
MobileWorld-Diffusion  & --                & Open-source       & 20B   & \texttt{xwk123/MobileWorld-Diffusion}      & \citep{xu2026mobileworldmodelguides} \\
\bottomrule
\end{tabular}%
}
\caption{World models evaluated in our benchmark.}
\label{tab:model_list}
\end{table*}
\newpage
\section{Case Study}
\label{app:case_study}

\begin{figure*}[t]
\centering
\includegraphics[width=\linewidth]{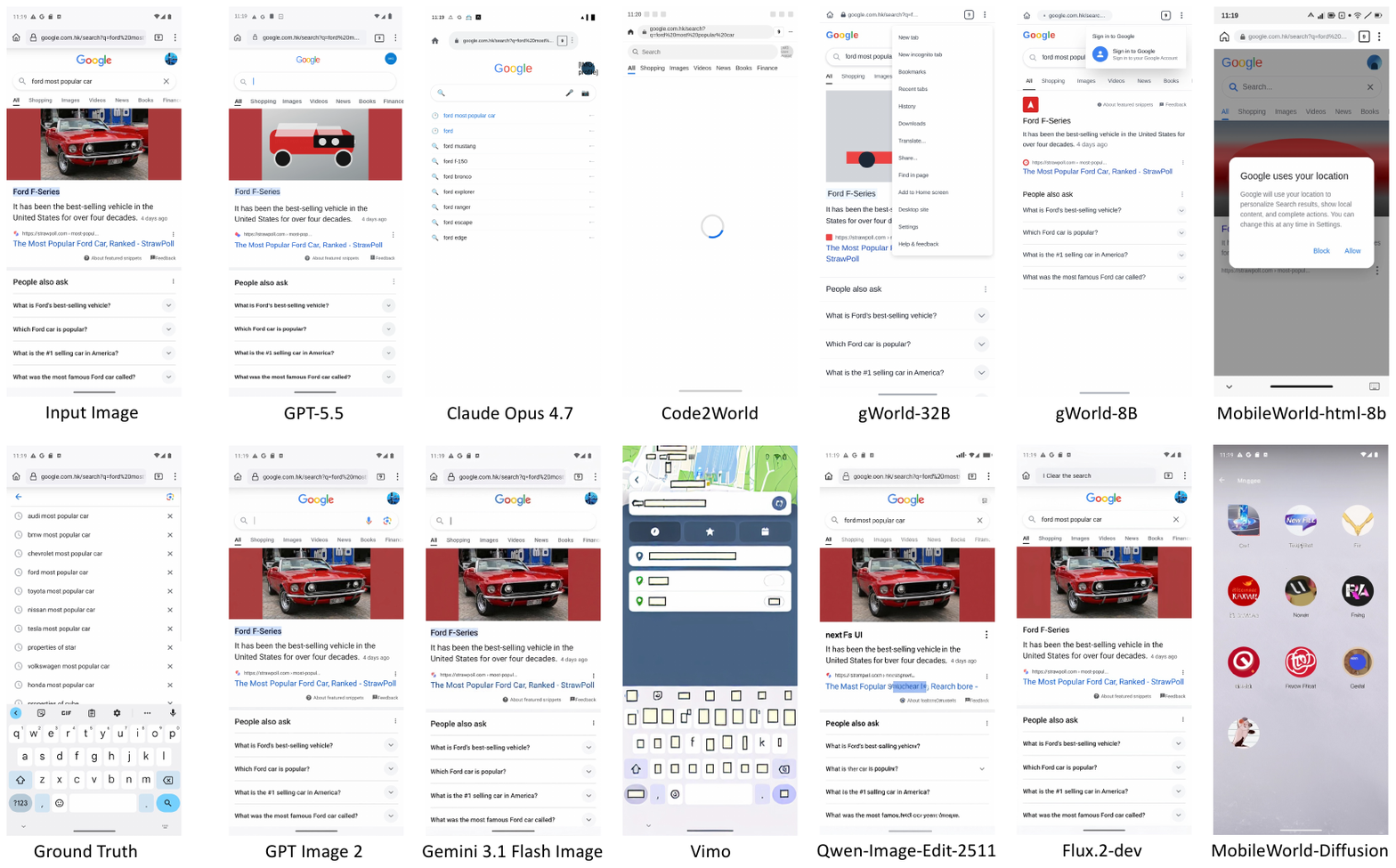}

\caption{
Single-step qualitative comparison for the action \texttt{Clear the search bar}.
The figure compares the predicted next UI states from different GUI world models
against the ground-truth next state.
}
\label{fig:single-step-case-clear-search}
\end{figure*}

\begin{figure*}[t]
\centering
\includegraphics[width=\linewidth]{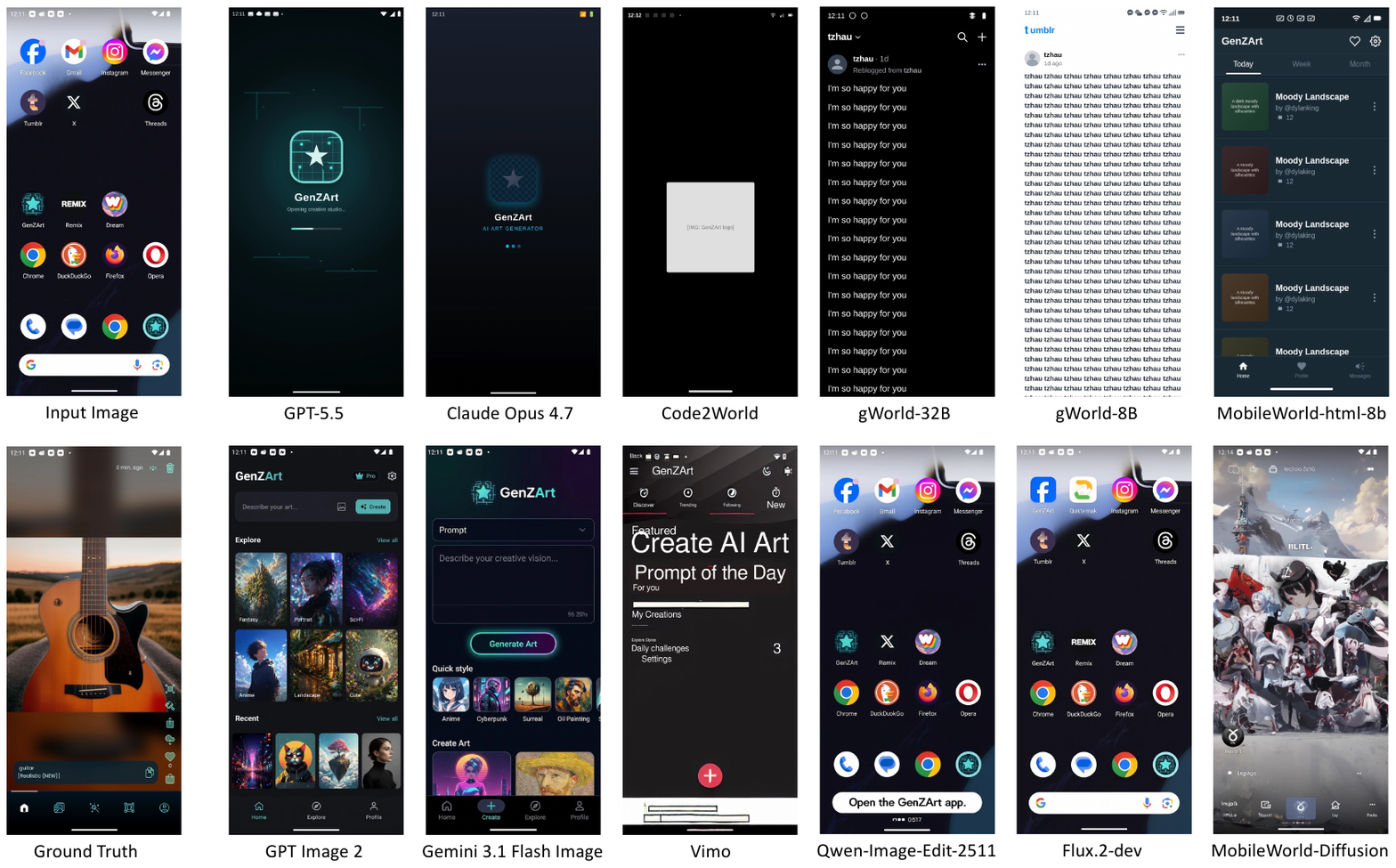}

\caption{
Single-step qualitative comparison for the action \texttt{Open the GenZArt app}.
The figure compares the predicted next UI states from different GUI world models
against the ground-truth next state.
}
\label{fig:single-step-case-open-app}
\end{figure*}

\begin{figure*}[t]
    \centering
    \captionsetup[subfigure]{skip=2pt}
    \setlength{\abovecaptionskip}{3pt}
    \setlength{\belowcaptionskip}{0pt}

    \begin{minipage}{0.84\textwidth}
        \centering
        \scriptsize
        \textbf{Task:}
        Search for ``vintage camera'', choose any listing from the results, add it to the Watchlist so the heart/watch icon becomes filled, then go to My eBay $\rightarrow$ Watchlist and confirm the listing appears there.
    \end{minipage}

    \vspace{0.2em}

    \begin{subfigure}{0.98\textwidth}
        \centering
        \includegraphics[
            width=\linewidth,
            trim={0 35 0 25},
            clip
        ]{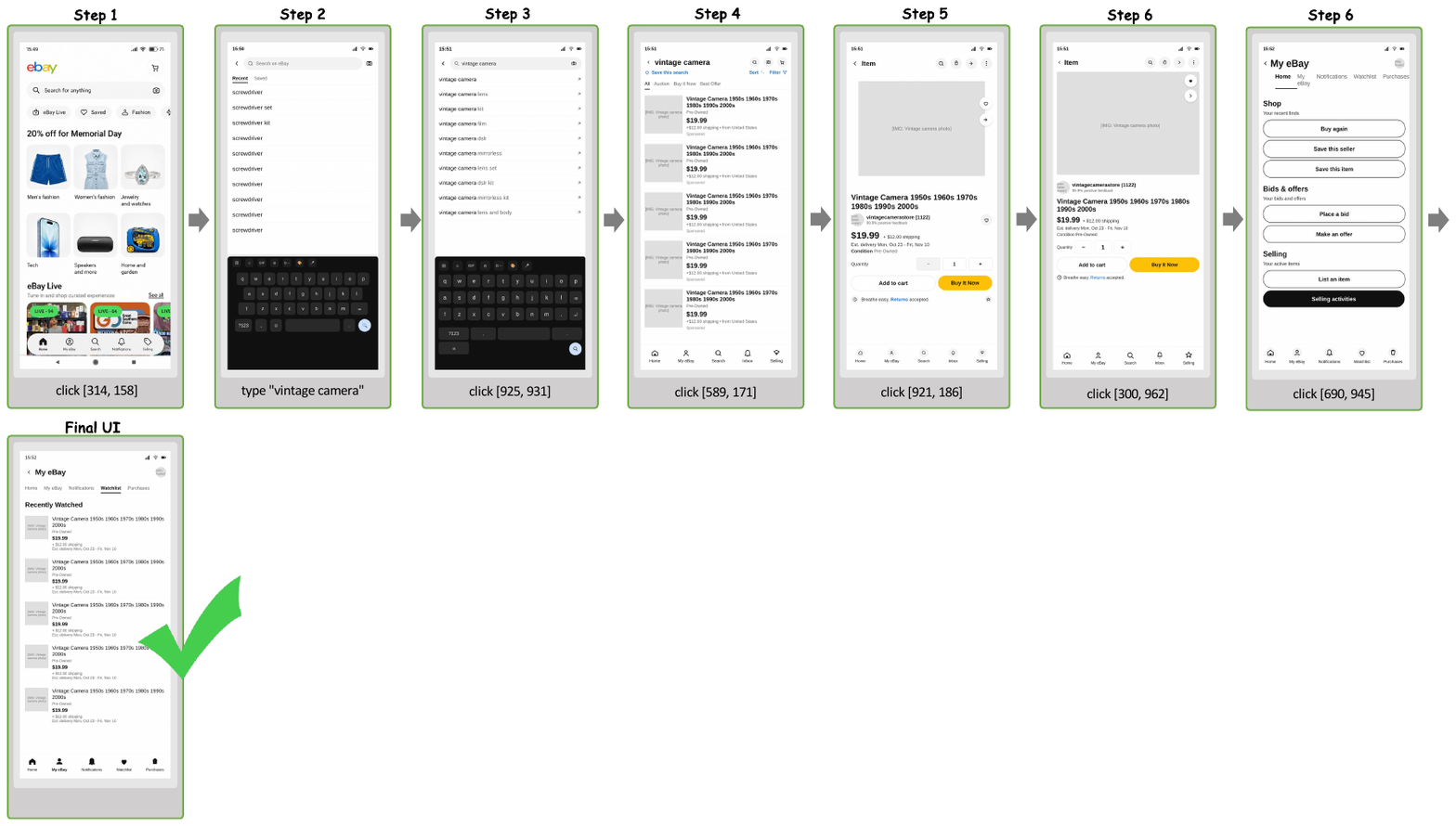}
        \caption{\textbf{With history.} The model maintains multi-step contextual consistency and successfully reaches the target final UI.}
        \label{fig:case_with_history_success}
    \end{subfigure}

    \vspace{-0.2em}

    \begin{subfigure}{0.98\textwidth}
        \centering
        \includegraphics[
            width=\linewidth,
            trim={0 35 0 25},
            clip
        ]{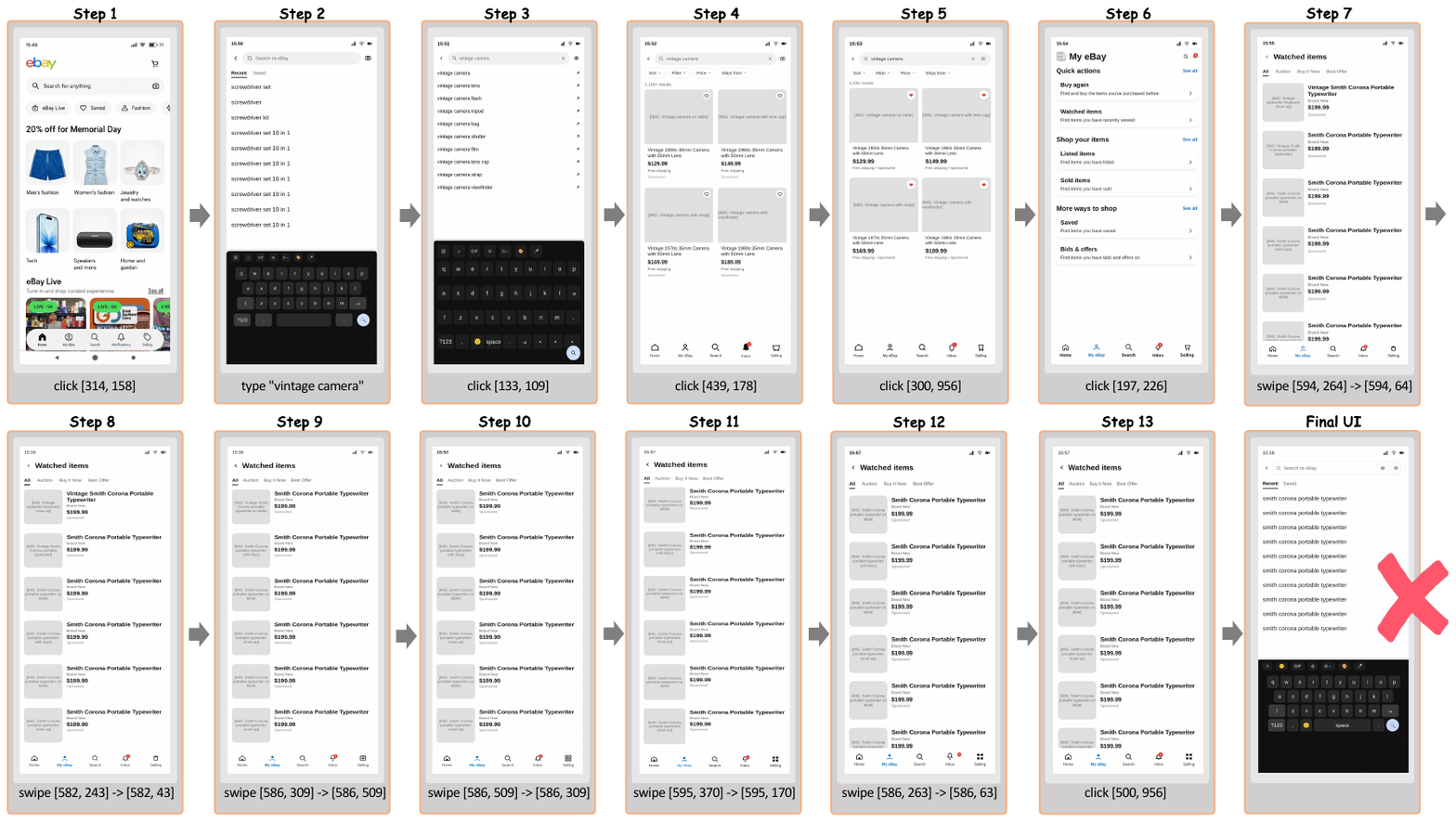}
        \caption{\textbf{Without history.} The model loses long-horizon contextual consistency, leading to an incorrect trajectory and a failed final UI.}
        \label{fig:case_without_history_fail}
    \end{subfigure}

    \vspace{-0.3em}

    \caption{
    Qualitative comparison between GUI world model (Code2World) rollouts with and without historical context.
    The history-conditioned rollout preserves the task state across multiple steps and successfully reaches the expected final UI,
    whereas the no-history rollout deviates from the intended trajectory and fails to complete the task.
    This example illustrates the importance of historical conditioning for long-horizon GUI state prediction and task-progress consistency.
    }
    \label{fig:appendix_history_case_study}
\end{figure*}

\begin{figure*}[t]
    \centering
    \setlength{\abovecaptionskip}{3pt}
    \setlength{\belowcaptionskip}{0pt}

    \begin{minipage}{0.82\textwidth}
        \centering
        \scriptsize
        \textbf{Task:}
        Using Yahoo Sports, find the date and time of the next NBA game.
        Once you have this information, create a reminder in your Calendar app.
    \end{minipage}

    \vspace{0.15em}

    {\small
    \textbf{Border annotation:}
    \successframe~successful / consistent transition;
    \failframe~failed / inconsistent transition.
    }

    \vspace{0.35em}

    \begin{subfigure}{0.96\textwidth}
        \centering
        \includegraphics[
            width=0.74\linewidth,
            trim={0 40 0 30},
            clip
        ]{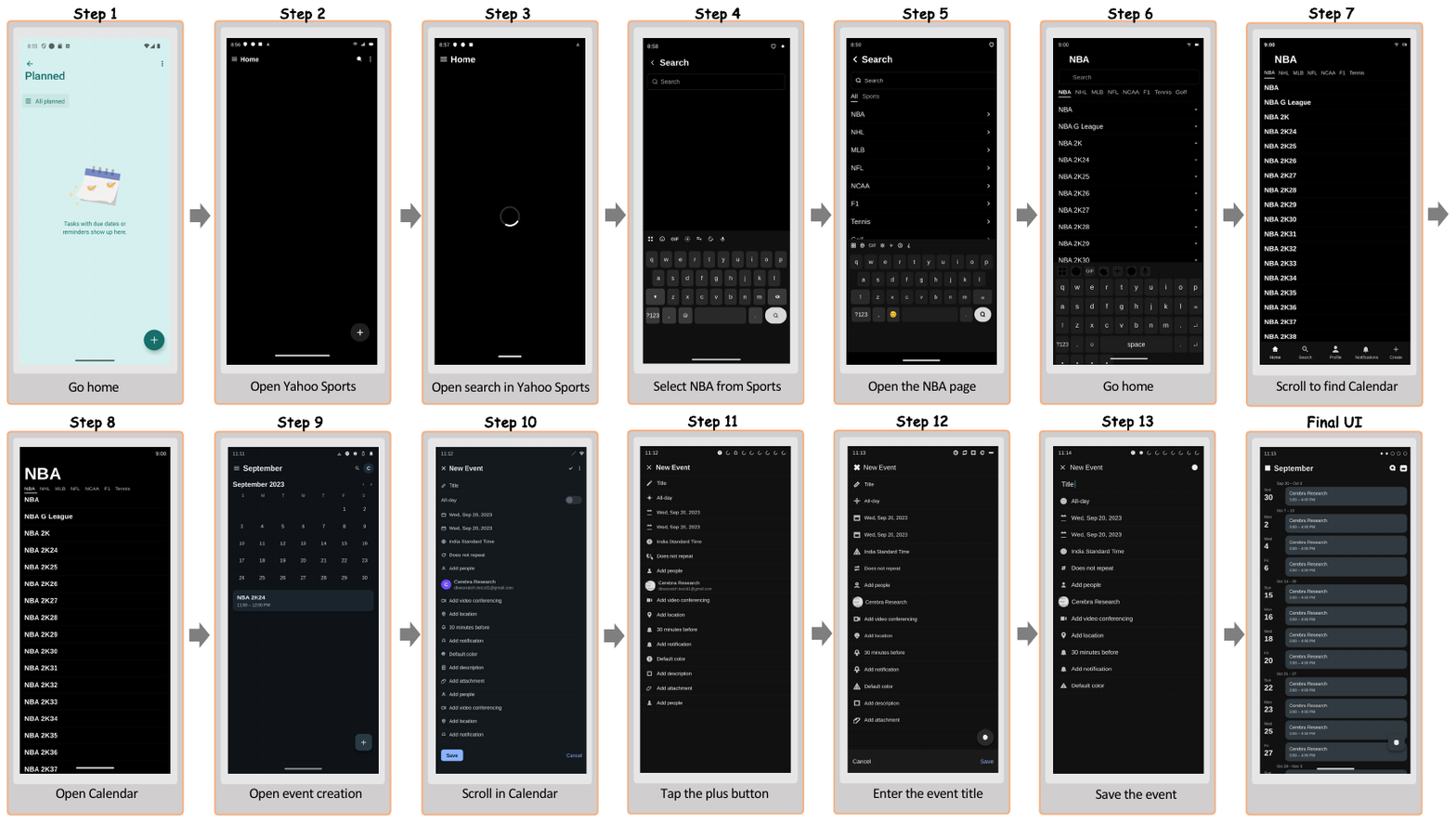}
        \caption{\textbf{Code2World.} Most transitions fail.}
        \label{fig:offline_case_code2world}
    \end{subfigure}

    \vspace{-0.65em}

    \begin{subfigure}{0.96\textwidth}
        \centering
        \includegraphics[
            width=0.74\linewidth,
            trim={0 40 0 30},
            clip
        ]{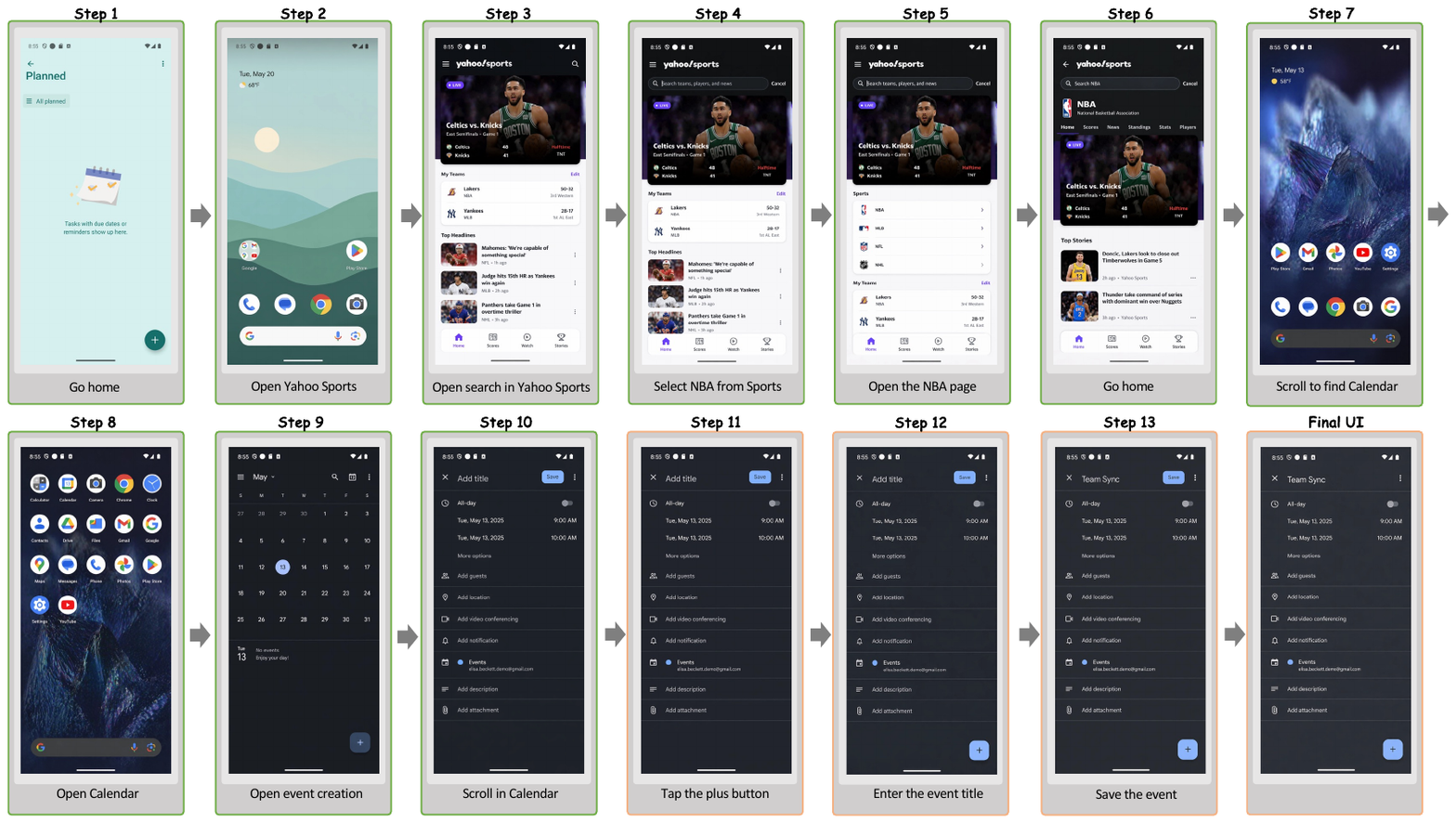}
        \caption{\textbf{GPT Image 2.} Early steps succeed, but later steps drift.}
        \label{fig:offline_case_gpt}
    \end{subfigure}

    \vspace{-0.65em}

    \begin{subfigure}{0.96\textwidth}
        \centering
        \includegraphics[
            width=0.74\linewidth,
            trim={0 40 0 30},
            clip
        ]{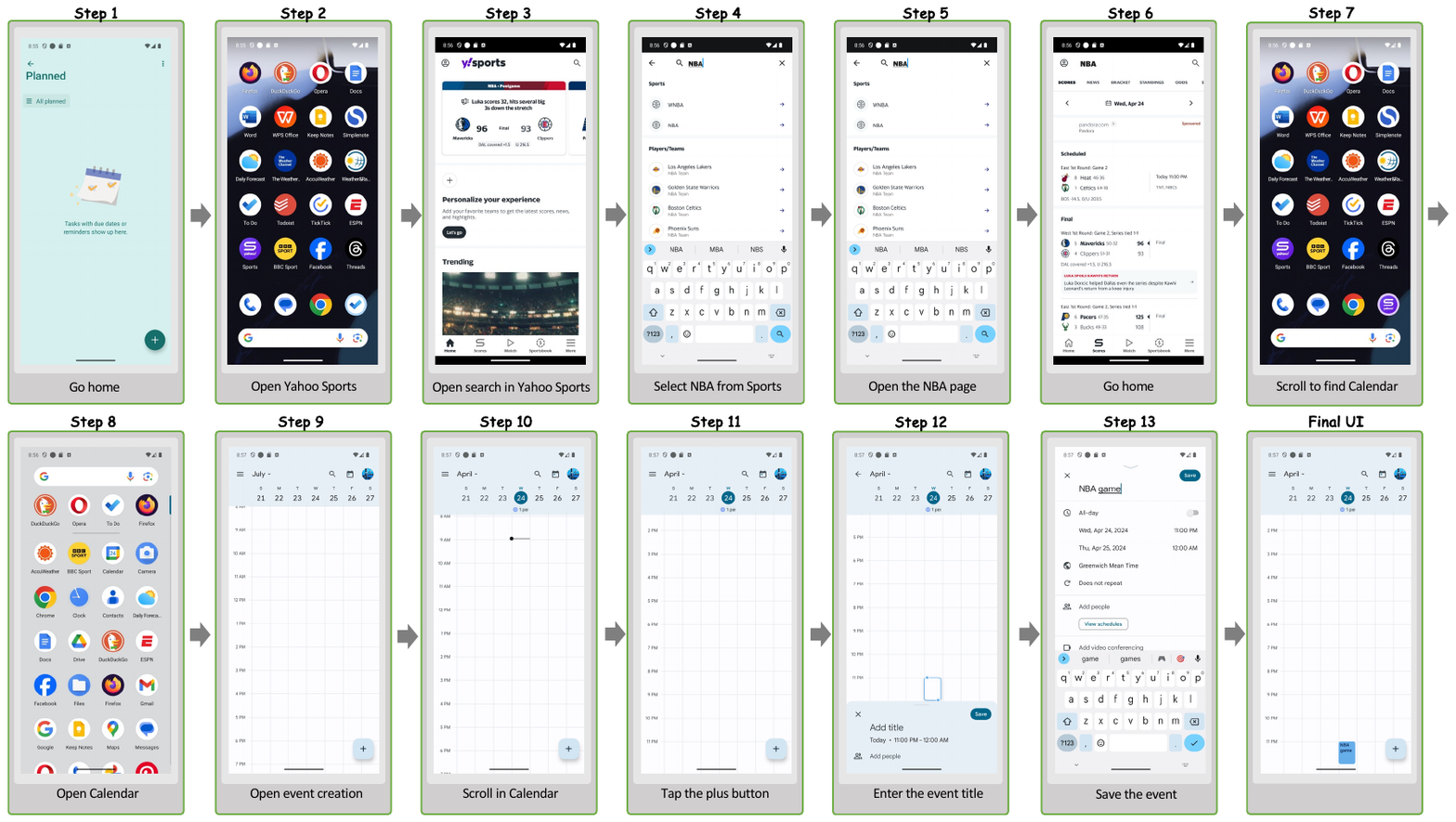}
        \caption{\textbf{Ground truth.} Reference trajectory.}
        \label{fig:offline_case_gt}
    \end{subfigure}

    \vspace{-0.65em}

    \caption{
    Offline rollout comparison under the same long-horizon cross-app action sequence.
    Green and orange borders indicate successful and failed step-level transitions, respectively.
    Code2World fails in most steps, while GPT Image 2 preserves early states but drifts later.
    }

    \label{fig:appendix_offline_rollout_comparison}
\end{figure*}
\newpage
\section{Offline Data Construction Details}
\label{app:data_construction}

This appendix provides the concrete rules behind the offline trajectory pipeline described in \autoref{sec:offline_track}.

\subsection{Filtering and Normalization}

Stage~1 applies several heuristic rules to remove low-quality trajectories from GUIOdyssey. We first exclude failed or interrupted episodes, keeping only trajectories whose final action marks a successful task completion. We then restrict trajectory length to between 5 and 20 steps, as shorter episodes provide little signal for evaluating long-horizon consistency while longer ones introduce uncontrolled complexity. To remove trajectories dominated by meaningless operations, we further require that the combined share of \texttt{tap}, \texttt{type}, and \texttt{long\_press} actions exceeds 50\% of all steps, that no action is repeated more than three times consecutively, and that launcher operations such as \texttt{HOME} and app switching account for at most 50\% of all steps. Finally, the recording device must be a phone, with tablet and foldable trajectories excluded due to their differing screen geometry. After filtering, we restore the normalized coordinates in the original dataset to pixel level, so that all benchmarked models can consume them under a common input convention.

\subsection{Single-Step Plausibility Verification}

Stage~2 catches transitions whose outcome cannot plausibly result from the recorded action alone. Typical failure modes include tapping a search bar and jumping directly to a result page while skipping text entry and submission, or tapping an app icon and landing on a deep in-app page rather than the entry screen. A trajectory is kept only if all of its transitions pass the check; any single flagged transition discards the full trajectory.

\subsection{Strongly State-Dependent Steps}

For the quality signal in Stage~3, a step is labeled \emph{strongly state-dependent} if its correct execution depends on information established by multiple earlier states rather than only the immediately preceding screen.
\end{document}